\documentclass[letterpaper]{article} % DO NOT CHANGE THIS
\usepackage{aaai24}  % DO NOT CHANGE THIS
\usepackage{times}  % DO NOT CHANGE THIS
\usepackage{helvet}  % DO NOT CHANGE THIS
\usepackage{courier}  % DO NOT CHANGE THIS
\usepackage[hyphens]{url}  % DO NOT CHANGE THIS
\usepackage{graphicx} % DO NOT CHANGE THIS
\usepackage{natbib}  % DO NOT CHANGE THIS AND DO NOT ADD ANY OPTIONS TO IT
\usepackage{caption} % DO NOT CHANGE THIS AND DO NOT ADD ANY OPTIONS TO IT
\usepackage{amsmath, amssymb, amsfonts}
\usepackage[linesnumbered, ruled, vlined]{algorithm2e}
\SetAlFnt{\small}
\SetAlCapFnt{\small}
\SetAlCapNameFnt{\small}
\SetArgSty{textrm}
\SetInd{0.1em}{0.5em}

\usepackage{etoolbox}
\makeatletter
\patchcmd{\@algocf@start}% <cmd>
  {-1.5em}% <search>
  {0pt}% <replace>
  {}{}% <success><failure>
\makeatother

\usepackage{algpseudocode}
\usepackage{array}
\usepackage{multirow}
\usepackage{diagbox}
\usepackage[dvipsnames]{xcolor}
\usepackage[title]{appendix}

\usepackage[shortlabels]{enumitem}

\DeclareMathOperator*{\argmax}{arg\,max\,}
\DeclareMathOperator*{\minimize}{minimize}

\usepackage{amsthm}
\usepackage{thmtools}
\declaretheoremstyle[%
  spaceabove=0pt,%reduce or increase between theorem and proof
  spacebelow=0pt,%reduce or increase
  headfont=\normalfont\itshape,%
  postheadspace=0.5em,%
  qed=\qedsymbol%
]{mystyle}

\renewcommand\qedsymbol{$\blacksquare$}

\newcommand{\primarykeywords}{
Multi-Agent Planning; Robotics
}

\usepackage[switch, modulo]{lineno}
\title{Multi-Robot Connected Fermat Spiral Coverage}
\author{
    Jingtao Tang, Hang Ma
}
\affiliations{
    Simon Fraser University\\
    \{jingtao\_tang, hangma\}@sfu.ca
}

\usepackage{bibentry}
\begin{document}
\nolinenumbers

\maketitle

\begin{abstract}
We introduce the Multi-Robot Connected Fermat Spiral (MCFS), a novel algorithmic framework for Multi-Robot Coverage Path Planning (MCPP) that adapts Connected Fermat Spiral (CFS) from the computer graphics community to multi-robot coordination for the first time. MCFS uniquely enables the orchestration of multiple robots to generate coverage paths that contour around arbitrarily shaped obstacles, a feature that is notably lacking in traditional methods. Our framework not only enhances area coverage and optimizes task performance, particularly in terms of makespan, for workspaces rich in irregular obstacles but also addresses the challenges of path continuity and curvature critical for non-holonomic robots by generating smooth paths without decomposing the workspace. MCFS solves MCPP by constructing a graph of isolines and transforming MCPP into a combinatorial optimization problem, aiming to minimize the makespan while covering all vertices. 
Our contributions include developing a unified CFS version for scalable and adaptable MCPP, extending it to MCPP with novel optimization techniques for cost reduction and path continuity and smoothness, and demonstrating through extensive experiments that MCFS outperforms existing MCPP methods in makespan, path curvature, coverage ratio, and overlapping ratio. Our research marks a significant step in MCPP, showcasing the fusion of computer graphics and automated planning principles to advance the capabilities of multi-robot systems in complex environments. Our code is publicly available at \protect\url{https://github.com/reso1/MCFS}.
\end{abstract}

\section{Introduction}
In the evolving landscape of multi-robot systems, the efficiency and effectiveness of Multi-Robot Coverage Path Planning (MCPP)~\cite{almadhoun2019survey} remain pivotal in a myriad of applications, ranging from environmental monitoring~\cite{collins2021scalable} to search-and-rescue operations~\cite{song2022multi} in complex workspaces. Traditional methodologies, such as cellular decomposition~\cite{latombe1991exact,acar2002morse} and grid-based methods~\cite{gabriely2001spanning,hazon2005redundancy}, have laid a solid foundation for understanding and navigating the challenges inherent in these tasks. However, as the complexity of environments and the demand for more efficient coverage increase, there is a growing need for innovative strategies that can adeptly handle workspaces rich in irregular obstacles with both high precision and adaptability.

This paper introduces a novel algorithmic framework, called Multi-Robot Connected Fermat Spiral (MCFS), which revolutionizes MCPP by building upon the principles of Connected Fermat Spiral (CFS)~\cite{zhao2016connected} from the computer graphics community. This represents the first application of leveraging CFS to solve MCPP challenges in automated planning and robotics, showcasing a unique interdisciplinary fusion. MCFS stands out for its unique ability to coordinate the robots in generating contour-like coverage paths, elegantly adapting to the intricacies of arbitrary-shaped obstacles---a characteristic not typically addressed by traditional methods. Its contouring ability also enhances task efficiency in both time and operation cost (e.g., energy) by balancing the path costs across multiple robots, as indicated by the makespan~\cite{zheng2010multirobot}. 

Besides task efficiency, a key challenge in MCPP is managing the deceleration and sharp turns required by nonholonomic robots. Traditional methods~\cite{lu2023tmstc, vandermeulen2019turn}, often focused on minimizing path turns, are restricted to rectilinear workspaces and rely on decomposing the area into rectangles. This approach is less effective in arbitrary-shaped environments. On the contrary, the essence of our MCFS framework lies in its global coverage strategy, conceptualizing the paths as a series of interconnected spirals that seamlessly integrate the movements of multiple robots. This strategy results in smooth covering paths without the need for decomposition, inherently accounting for path curvature---a vital factor for efficient robotic navigation.

Drawing inspiration from the original application of CFS in additive manufacturing~\cite{gibson2021additive}, our MCFS framework innovatively adapts CFS to tackle the MCPP problem, which generates continuous and smooth coverage paths by converting a set of equidistant contour-parallel isolines into connected Fermat spirals. MCFS first constructs a graph of isolines, associating each vertex with an isoline and connecting it to associated vertices of adjacent isolines. It then reduces the MCPP problem to Min-Max Rooted Tree Cover (MMRTC), a combinatorial optimization problem that finds a forest of trees to cover all vertices of the graph while minimizing the makespan. Our framework is versatile, allowing coverage paths to start from arbitrary starting points as required in MCPP, and optimizes the distribution of the coverage of both multiple whole isolines and segments of an isoline among multiple robots, showcasing an innovative approach to effectively managing the makespan, curvature, and path continuity for each robot.

We conclude \textbf{our key contributions} as follows: 
(1) We propose a unified version of CFS that standardizes the stitching of adjacent isolines, allowing for customized priorities in selecting stitching points and providing scalability and ease of adaptation to MCPP by enabling coverage paths to start from any given initial robot positions. (2) We demonstrate how our MCFS extends this unified version of CFS to MCPP and effectively solves the corresponding MMRTC problem. (3) We introduce two optimization techniques: one that adds edges between non-adjacent but connectable pairs of isolines to expand the solution space and another that refines the MMRTC solution for balanced path costs and reduced overlap in multi-robot coverage. (4) We present extensive experimental results validating the superiority of our MCFS over state-of-the-art MCPP methods in metrics of makespan, path curvature, coverage ratio, and overlapping ratio, showcasing its effectiveness in diverse coverage scenarios.

\section{Related Work}
We categorize existing Single-Robot Coverage Path Planning (CPP) and MCPP methods into grid-based, cellular decomposition, and global methods.
We refer interested readers to \cite{Tomaszewski-2020-125840} for a more detailed taxonomy.

\noindent\textbf{Grid-Based Methods:}
Grid-based coverage methods abstract workspaces into square grids~\cite{hazon2005redundancy,kapoutsis2017darp, tang2021mstc}, allowing for the application of various graph algorithms.
One prominent method, Spanning Tree Coverage (STC)~\cite{gabriely2001spanning}, constructs a minimum spanning tree and then generates circumnavigating paths on the tree to cover the workspace. 
STC-based MCPP methods~\cite{hazon2005redundancy,tang2023mixed,tang2024large} work by finding a set of trees that jointly visit all vertices and assigning each robot a path that circumnavigates a tree. While convenient, the complexity of optimally solving grid-based MCPP grows exponentially in the workspace size and the number of robots.

\noindent\textbf{Cellular Decomposition Methods:}
These methods decompose the workspace into sub-regions by detecting geometric critical points, such as trapezoid~\cite{latombe1991exact} and Morse~\cite{acar2002morse} decomposition.
CPP methods generate zigzag paths in these subregions for coverage~\cite{choset2000coverage,wong2003topological}, and MCPP methods connect and assign these subregions, filled with zigzag paths, to robots for cooperative coverage~\cite{rekleitis2008efficient,mannadiar2010optimal,karapetyan2017efficient}.
Additionally, some research optimizes the direction of the zigzag paths for single robots~\cite{oksanen2009coverage,bochkarev2016minimizing}.
Although efficient, these methods are less suitable for obstacle-rich or nonrectilinear workspaces due to their reliance on geometric partitioning.

\noindent\textbf{Global Methods:}
Global CPP methods directly generate paths to cover the workspace without decomposing it.
They fall into two types: the first type generates separate paths that contour around obstacles~\cite{yang2002equidistant}, and the second type generates a closed path, including Spiral Path~\cite{ren2009combined} and CFS that are notable for their continuous and smooth paths. CFS paths are especially convenient as their entry and exit points are adjacent, facilitating the integration of multiple paths.
A recent paper has built a CFS path based on an exact geodesic distance field to cover a terrain surface~\cite{wu2019energy}. However, to our knowledge, there are no global methods yet developed for MCPP.

\section{Connected Fermat Spiral (CFS)}~\label{sec:cfs}
In this section, we present our unified version of CFS, an adaptation of the original CFS concept.
The original CFS employs a two-phase process to transform a set of equidistant isolines into a closed path that covers an input polygon workspace. It utilizes a graph structure, where vertices represent individual isolines and edges connect vertices whose respective isolines have adjacent segments.
%For simplicity, we assume that the exit point is one point before the entry point in counterclockwise order.
Initially, the original CFS identifies a set of ``pockets''---connected components on the spanning tree of the graph. The first phase transforms the isolines within each pocket into a \textit{Fermat spiral}~\cite{lockwood1967book}, and the second phase stitches these isolated Fermat spirals to construct the final, connected Fermat spiral by traversing the pockets using the graph edges.
The details of the original CFS can be found in Appendix A.

Our unified version of CFS modifies the graph construction of isolines and consolidates the original two-phase process into a singular, cohesive operation for the CPP problem. The primary modification in our approach lies in the stitching phase. 
Rather than explicitly identifying pockets and then stitching the resulting isolated Fermat spirals, our method integrates a unified process that simultaneously addresses both the conversion of isolines within a pocket into Fermat spirals and the interconnection of these spirals. This integrated process is applied to every stitchable pair of isolines, effectively merging the conversion and stitching phases. By traversing a rooted spanning tree of the graph, the same connected Fermat spiral as the original CFS is obtained.
The advantage of our unified CFS approach is twofold. Firstly, it enhances scalability, facilitating the incorporation of diverse utilities within the framework. Secondly, it simplifies the extension of CFS to MCPP.
%By integrating and streamlining the conversion and stitching processes, our approach improves the efficiency and versatility of CFS for complex CPP challenges.

\subsection{Constructing Isolines and the Isograph}\label{subsec:isograph}
We describe our approach for generating layered isolines from a given polygon workspace to be covered and building the isograph. The polygon is enclosed by its boundary, consisting a set of interior boundary polylines that represent obstacles and an exterior boundary polyline.

\noindent\textbf{Generating Layered Isolines:}
The procedure starts by uniformly sampling a 2D mesh grid of points within the polygon. A distance field is built for these points, representing their shortest distance to the polygon boundary (encompassing both the interior obstacle boundary polylines and the exterior boundary polyline). We denote the distance between isolines at adjacent layers as $l$, and the largest distance to the polygon boundary among all points as $l_{max}$. We then use the \textit{Marching Squares} algorithm~\cite{maple2003geometric} to generate layered isolines for each layer $i=1, 2, ..., \lfloor l_{max}/l \rfloor$. This ensures that the distance between each point in the layer-$i$ isoline and the polygon boundary is $l\times i$. The last step resamples equidistant points along each isoline, maintaining a consistent distance of $l$ between adjacent points.

\begin{figure}[tb]
\centering
\includegraphics[width=0.8\columnwidth]{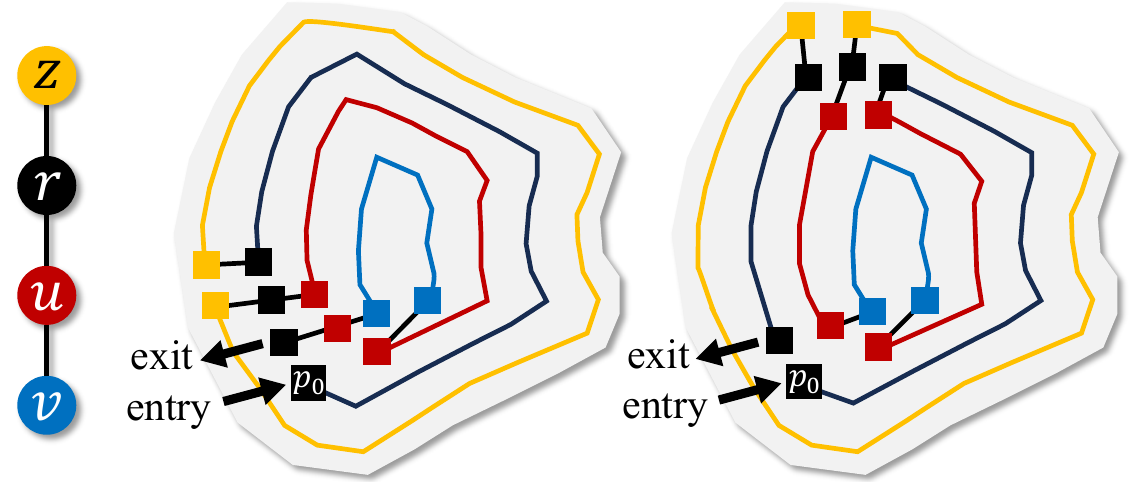}
\caption{The unified version of CFS on a workspace (grey region). Colored squares represent the stitching tuples. From left to right: The input isograph, the path resulting from the CFS selector, and the path resulting from the MCS selector.}
\label{fig:uCFS_demo}
\end{figure}

\noindent\textbf{Building the Isograph:}
We define \textit{isograph} of the layered isolines as an undirected graph $G=(V,E)$, where $V$ is the set of \textit{isovertices}, each associated with a unique isoline.
For ease of reference, we let $I_v$ and $L_v$ denote the isoline associated with any $v\in V$ and its respective layer. Similar to the original CFS, we define a \textit{connecting segment set} $O_{u\rightarrow v}$ for any pair of isovertices $u, v\in V$ in adjacent layers (i.e., $|L_u - L_v| = 1$) as:
\begin{align}\label{eqn:O_u2v}
\resizebox{0.9\linewidth}{!}{
$O_{u\rightarrow v}=\{\mathbf{p}\in I_u\,|\,\forall z\in V, d(\mathbf{p},I_v)<d(\mathbf{p}, I_z) \wedge L_z=L_v\}$}
\end{align}
where $d(\mathbf{p}, I)$ denotes the distance between point $\mathbf{p}$ and isoline $I$.
Unlike the original CFS which directly constructs an undirected edge $(u, v)$ if $O_{u\rightarrow v}$ is nonempty, we also consider $O_{v\rightarrow u}$ for edge construction. This consideration provides flexibility in traversing the isograph in any order and from any root isovertex in the CFS context.
It also avoids adding edges $(u,v)$ where the respective isolines $I_u$ and $I_v$ are separated by multiple isolines, as such pairs may be unsuitable for stitching in the CPP context (see Fig.~\ref{fig:case_study_cfs} for the case study). Therefore, we define a set $O_{u,v}$ of \textit{stitching tuples} for any $u, v\in V$ in adjacent layers as:
\begin{align}\label{eqn:O_uv}
\resizebox{0.9\linewidth}{!}{
$O_{u,v}=\{(\mathbf{p},\mathbf{q})\in O_{u\rightarrow v}\times O_{v\rightarrow u}\,|\, \mathbf{p}=\mathcal{C}_u(\mathbf{q})\wedge \mathbf{q}=\mathcal{C}_v(\mathbf{p})\}$}
\end{align}
where $\mathcal{C}_u(\mathbf{p})$ denotes the nearest point along isoline $I_u$ to point $\mathbf{p}$.
Subsequently, an undirected edge $(u,v)$ is formed for any $u,v\in V$ in adjacent layers with a nonempty $O_{u,v}$.
Each $(\mathbf{p},\mathbf{q})\in O_{u,v}$ serves as a candidate stitching tuple to connect isolines $I_u$ and $I_v$ by stitching $\mathbf{p}$ to $\mathbf{q}$ and $\mathcal{B}_u(\mathbf{p})$ to $\mathcal{B}_v(\mathbf{q})$, where $\mathcal{B}_u(\mathbf{p})$ denotes the point preceding $\mathbf{p}$ along isoline $I_u$ in counterclockwise order.
Fig.~\ref{fig:uCFS_demo} shows how four isolines are connected via the squares as the stitching points.

Although the original CFS assigns a weight of $|O_{u\rightarrow v}|$ to each edge to retain a low-curvature path when determining the isograph traversal order for connecting isolated Fermat spirals, we currently leave the edge weight definition application-specific and will explicitly address this objective for every stitching operation in the stitching tuple selector.

\begin{algorithm}[tb]
\linespread{0.6}\selectfont
\DontPrintSemicolon
\caption{Unified Version of CFS}\label{alg:cfs}
\SetKwInput{KwInput}{Input}
\KwInput{isograph $G$, entry point $\mathbf{p}_0$}
$r\gets$ the isovertex of $G$ containing $\mathbf{p}_0$\;\label{alg.cfs.identify_root}
$\pi\gets I_r(\mathbf{p}_0),\; U\gets\emptyset$\;\label{alg.cfs.init_2}
\For{$(u,v)\in$ DFS traversal edges of $G$ from $r$}{\label{alg.cfs.loop}
    remove any $(\mathbf{p},\mathbf{q})$ from $O_{u,v}$ where $\mathbf{p}\in U$ or $\mathbf{q}\in U$\;\label{alg.cfs.filter}
    $(\mathbf{p}, \mathbf{q}) \gets f\left(O_{u,v}\right)$\Comment{by any stitching tuple selector $f$}\;\label{alg.cfs.select}
    stitch $I_v(\mathbf{p})$ into $\pi$ by stitching $\mathbf{p}$ to $\mathbf{q}$ and $\mathcal{B}(\mathbf{p})$ to $\mathcal{B}(\mathbf{q})$\;\label{alg.cfs.stitch}
    $U\gets U\cup\{\mathbf{p}, \mathbf{q}\}$\;\label{alg.cfs.update}
}
\Return $\pi$\;
\end{algorithm}

\subsection{Unifying the CFS Algorithm}
We detail our unified version of CFS in Alg.~\ref{alg:cfs}, which takes as input an isograph $G$ and an entry point $\mathbf{p}_0$. The algorithm starts by identifying the isovertex $r$ containing $\mathbf{p}_0$ [Line~\ref{alg.cfs.identify_root}] as the root for a depth-first search (DFS) traversal of $G$.
It then initializes the CFS path $\pi$ to be constructed and the set $U$ to record the points already used to stitch the isolines [Line~\ref{alg.cfs.init_2}].
The main loop then iterates over the DFS edges [Line~\ref{alg.cfs.loop}] and stitches the corresponding pair of isolines for each edge [Lines~\ref{alg.cfs.select}-\ref{alg.cfs.stitch}].
Specifically, a stitching tuple $(\mathbf{p}, \mathbf{q})$ is selected via any selector [Line~\ref{alg.cfs.select}].
For any isovertex $v\in V$, we use $I_v(\mathbf{p})$ to denote the counterclockwise path along isoline $I_v$ starting at $\mathbf{p}$ and ending at $\mathcal{B}_v(\mathbf{p})$. This path segment is then stitched into $\pi$ using the selected stitching tuple [Line~\ref{alg.cfs.stitch}].
The set $U$ is updated to include these newly selected stitching tuples [Line~\ref{alg.cfs.update}].
Following the iterations over all DFS edges, the final path $\pi$ is constructed to stitch together all isolines and completely cover the given polygon.

\subsection{Stitching Tuple Selector}\label{subsec:selector}
We now propose three stitching tuple selectors, each designed to select an appropriate stitching tuple $o$ from a given set $O_{u,v}$ for connecting isolines $I_u$ and $I_v$.
Fig.~\ref{fig:uCFS_demo} demonstrates an example of these selectors.
% \begin{figure}
% \centering
% \includegraphics[width=0.85\columnwidth]{figs/stitching_pts_candidate.png}
% \caption{The four candidate stitching tuples for any $(\mathbf{p},\mathbf{q})\in O_{u,v}$ to stitch isolines $I_u$ and $I_v$ (black solid line). Each candidate stitching tuple is represented by a pair of two dashed lines highlighted in the same color.}
% \label{fig:stitching_pts_candidate}
% \end{figure}

\noindent\textbf{Random Selector:}
The random selector $f_{\text{rnd}}$ randomly selects a stitching tuple from the set $O_{u,v}$.

\noindent\textbf{Connected Fermat Spiral (CFS) Selector:} 
The CFS selector $f_{\text{cfs}}$ aligns our unified version of CFS with the original CFS. It attempts to select a stitching tuple from $O_{u,v}$ for $(u,v)\in E$ that is adjacent to the previously selected stitching tuple of $(r, u)\in E$ or $(r, v)\in E$. Either $(r,u)$ or $(r,v)$, with its stitching tuple already selected by $f_{\text{cfs}}$, will be visited before $(u,v)$ in the DFS traversal (Line~\ref{alg.cfs.loop}). Assuming that $(r,u)$ is visited first with the selected stitching tuple $(\mathbf{p}', \mathbf{q}')\in O_{r,u}$, $f_{\text{cfs}}$ then checks for $o=(\mathbf{p}, \mathbf{q})$ in $O_{u,v}$ where $\mathcal{B}(\mathbf{p})=\mathbf{q}'$.
If such a tuple exists, it is selected for $(u,v)$; otherwise, the first tuple in $O_{u,v}$ is selected.

\noindent\textbf{Minimum Curvature Stitching (MCS) Selector:} The MCS selector $f_{\text{mcs}}$ iterates through $O_{u,v}$ to identify the stitching tuple $o=(\mathbf{p}, \mathbf{q})$ that minimizes the curvature difference $\Delta\kappa(o)$ before and after stitching, defined as:
\begin{align}\label{eqn:delta_kappa}
\Delta\kappa(o)=\sum\nolimits _{\mathbf{p}\in o} \left[\kappa_\pi(\mathbf{p})-\kappa_{I_u}(\mathbf{p})\right]
\end{align}
where $\kappa_\pi(\mathbf{p})$ and $\kappa_{I_u}(\mathbf{p})$ denote the curvatures at any point $\mathbf{p}$ on the new stitched path $\pi$ using $o$ and on the original isoline $I_u$, respectively.
Formally, the MCS selector is defined as $f_{\text{mcs}}(O_{u,v})=\argmax_{o\in O_{u,v}}\Delta\kappa(o)$.

\subsection{Case Study: Unified vs Original CFS}\label{app:CFS:case_study}
\begin{figure}[tb]
\centering
\includegraphics[width=0.995\columnwidth]{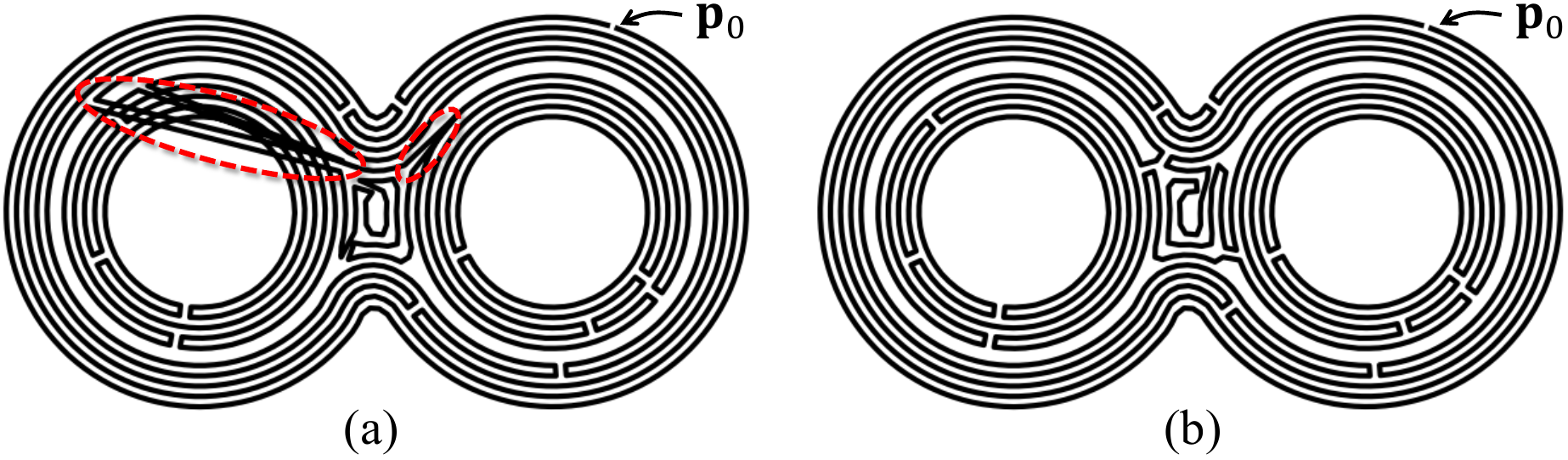}
\caption{The CFS paths for (a) the original unidirectional $O_{u\rightarrow v}$ in~\citet{zhao2016connected}, with the artifacts outlined in red dashed circles, and (b) our bidirectional $O_{u, v}$.}
\label{fig:case_study_cfs}
\end{figure}

We discuss the necessity of modification in the construction of the isograph edge set of our unified version of CFS in the CPP context. Unlike the original CFS that uses a unidirectional $O_{u\rightarrow v}$ in Eqn.~(\ref{eqn:O_u2v}) for edge set construction and always starts traversal from the lowest-layer isovertices, our unified CFS defines a more versatile bidirectional $O_{u,v}$ (Eqn.~(\ref{eqn:O_uv})). This modification addresses the requirement in CPP (and MCPP) for starting a coverage path from an arbitrary given point $\mathbf{p}_0$, as accommodated by Alg.~\ref{alg:cfs}. Our unified CFS starts the graph traversal from isovertex $r$, whose respective isoline contains $\mathbf{p}_0$, without the restriction of $r$ being the lowest-layer isovertex. Consequently, valid stitching tuples may not exist for edge construction if only single-directional tuples from layer $i$ to layer $i+1$ are considered as in the original CFS. 
Moreover, an isovertex $u$ with a local innermost isoline may find a nonempty $O_{u\rightarrow v}$ for any isovertex $v$ with $L_v=L_{u}+1$, recognizing $(u,v)$ as an edge, which potentially introduces path overlapping.
Fig.~\ref{fig:case_study_cfs}-(a) exemplifies such cases where some local innermost isolines are stitched to the isolines at adjacent layers yet separated by other isolines, a scenario effectively managed in our unified CFS (Fig.~\ref{fig:case_study_cfs}-(b)) but problematic in using the original CFS definitions.

% \begin{figure}[tb]
% \centering
% \includegraphics[width=0.8\columnwidth]{figs/selector_comp.pdf}
% \caption{CFS paths resulting from the (a) random selector, (b) CFS selector, (c) MCS selector, and (d) MCS selector with $O_{u\rightarrow v}$. Black triangles, blue lines, and red lines are the entry and exit points, the stitching path segments, and the removed isoline segments after stitching.}
% \label{fig:selector_comp}
% \end{figure}

%\subsection{Visualizing the Stitching Tuple Selectors}\label{app:CFS:comp_selectors}
% Fig.~\ref{fig:selector_comp} further visualizes the three stitching tuple selectors in Sec.~\ref{subsec:selector}. Fig.~\ref{fig:selector_comp}-(b) visualizes the staircase-like stitching scheme in the original CFS~\cite{zhao2016connected} using the CFS selector, while Fig.~\ref{fig:selector_comp}-(c) shows that the MCS selector always picks the stitching tuples at high-curvature positions in order to minimize the curvature.

\section{Multi-Robot CFS Coverage}\label{sec:mcfs}
In this section, we present our MCFS framework for solving MCPP. MCFS computes multiple trees from an input isograph, each corresponding to a different robot, and then applies CFS on each tree to compute individual coverage paths. 
First, we detail the CFS-based formulation of MCPP and introduce its reduction to Min-Max Rooted Tree Cover (MMRTC)~\cite{even2004min,tang2023mixed}. %Since there can still be unnecessary repetition in the coverage paths resulting from an optimal MMRTC solution,
We then present two optimization techniques, isograph augmentation and MMRTC solution refinement, aiming to further enhance the MCPP solution.

\subsection{Problem Formulation}\label{subsec:problem_formulation}
We present our problem formulation of MCPP that facilitates the extension of CFS. The problem of MCPP is to find a set $\Pi=\{\pi_i\}_{i\in I}$ of coverage paths for a set $I$ of robots that minimizes the makespan (i.e., the maximum path cost).
Following the existing literature~\cite{zheng2010multirobot, tang2021mstc}, we assume that each robot starts and ends at a given position, corresponding to a pair of adjacent entry and exit points in the CFS context.
Formally, the objective of MCPP is minimizing the makespan $\tau$, represented as:
\begin{align}\label{eqn:mcpp}
\min_{\Pi}\tau = \min_{\Pi=\{\pi_i\}_{i\in I}}\max\{c(\pi_1), c(\pi_2), ..., c(\pi_{|I|})\}.
\end{align}
When using CFS to generate each coverage path in $\Pi$, the path length is linear in $|\pi|$ and therefore the cost of any path $\pi$ can be evaluated as $c(\pi)=|\pi|$, since each isoline in CFS contains equidistant points (as detailed in the last section).
For an isograph $G=(V,E)$, each $v\in V$ is assigned a weight $w_v=|I_v|$, representing the number or points in isoline $I_v$. Consequently, the cost of any tree $T\subseteq G$ is $c(T)=\sum_{v\in V(T)} w_v$.
%It follows that we can extend CFS for MCPP by reducing it to the Min-Max Rooted Tree Cover (MMRTC) problem~\cite{even2004min,tang2023mixed}. 
The MMRTC problem parallels MCPP in its aim of finding a makespan-minimizing set of rooted trees, where each graph vertex is covered by at least one tree.
Given a graph $G=(V,E)$ and a set $R=\{r_i\}_{i\in I}\subseteq V$ of root isovertices for robots, the objective of MMRTC is defined as:
\begin{align}\label{eqn:mmrtc}
\min_{\mathcal{T}=\{T_i\}_{i\in I}} \max\{c(T_1), c(T_2), ..., c(T_{|I|})\}
\end{align}
where each $T_i\in\mathcal{T}$ is a tree rooted at $r_i$ and $c(T_i)$ is its tree cost.
Let $V(T)$ and $E(T)$ denote the vertex set and edge set of any tree $T$, respectively. The solution set $\mathcal{T}$ must satisfy $v\in\bigcup_{i\in I} V(T_i)$ to ensure the coverage of all $v\in V$.
Since the CFS stitches each isoline $I_v$ of $v\in V(T_i)$ to construct the coverage path $\pi_i\in\Pi$, we have $c(\pi_i)=|\pi_i|=\sum_{v\in V(T_i)}|I_v|={c(T_i)}$.
Therefore, for any isograph $G$ and the set $R$ of root isovertices for robots, the objective values in Eqn.~(\ref{eqn:mcpp}) and Eqn.~(\ref{eqn:mmrtc}) are identical under CFS, effectively reducing MCPP to MMRTC.

We employ the Mixed Integer Programming (MIP) model proposed in~\cite{tang2023mixed} to solve MMRTC optimally. The optimal set of trees obtained is then used to produce coverage paths by applying our unified CFS (Alg.~\ref{alg:cfs}) on each tree. Figs.~\ref{fig:pairwise_splitting}-(a) and (b) illustrate a 2-tree MMRTC instance and its corresponding solution. 
% For more details about the MIP model, see Appendix B in the full version of this paper~\cite{}. 
The details of the MIP model are shown in Appendix B. 

\subsection{Optimization: Isograph Augmentation}~\label{subsec:augmentation}
Recall that the isograph building process considers each edge only for two isolines in adjacent layers. This process, while efficient, often results in a sparse graph structure in the isograph and thus an undesirable MMRTC solution where certain isovertices are repetitively covered by multiple trees. One common example of such repetition appears for a \textit{cut isovertex}, defined as a vertex whose removal increases the number of connected components in the graph. Such repetitions become more common as the number of trees (robots) increases or when tree roots are clustered, thereby leading to increased makespan and reducing the overall quality of MCPP solutions.
To mitigate this issue, we propose to augment the sparse isograph with additional edges connecting isovertices in nonadjacent layers. This augmentation aims to reduce the sparsity of the isograph and allow MMRTC trees to explore new routes for joint coverage, thereby reducing repetitions and balancing tree costs.
% By reducing the isograph sparsity, the optimal MMRTC solution has no repetition on non-root isovertices as in Theorem~\ref{theo:repetition}, leading to a higher quality of the MCPP solution.
\begin{figure}[tb]
\centering
\includegraphics[width=0.75\columnwidth]{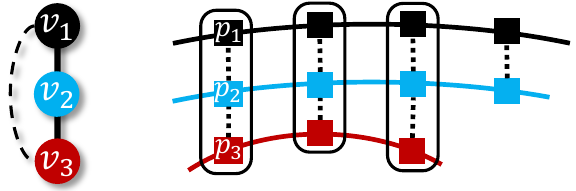}
\caption{Left: The augmented isograph with original edges (solid lines) and an augmented edge (dashed line). Right: Three sequences of stitching tuples (black boxes) for $O_{v_1,v_3}$.}
\label{fig:augmentation}
\end{figure}

The augmentation of an isograph $G=(V,E)$ operates by adding a set $E^\#$ of augmented edges, defined as:
\begin{align}\label{eqn:augmentation}
E^\# =\{(u,v)\,|\,\forall u,v\in V, 2\leq d_G(u,v)\leq\delta\}
\end{align}
where $d_G(\cdot,\cdot)$ denotes the graph distance between any two isovertices in $G$, and $\delta$ is a hyperparameter that sets the augmentation level.
For the edges in $E^\#$, stitching tuples are constructed differently from those edges in the original isograph edge set $E$. Without loss of generality, we consider an edge $(v_1,v_{k+1})\in E^\#$ and its shortest path $(v_1, v_2, ..., v_{k+1})$ in the original $G$ (i.e., each segment $(v_i, v_{i+1})$ is part of $E$ and $k$ is the graph distance between $v_1$ and $v_{k+1}$). The set $O_{v_1,v_{k+1}}$ comprises all pairs of $\mathbf{p}_1$ on the isoline of $v_1$ and $\mathbf{p}_{k+1}$ on the isoline of $v_{k+1}$ that can be feasibly connected, forming valid stitching tuples $(\mathbf{p}_1,\mathbf{p}_{k+1})$. Such points are connectable iff they form a sequence of consecutive stitching tuples $(\mathbf{p}_1, \mathbf{p}_2)\in O_{v_1, v_2}, \ldots, (\mathbf{p}_k, \mathbf{p}_{k+1})\in O_{v_k, v_{k+1}}$, which ensures that the straight-line segment between the pair does not intersect more than $k-1$ isoline(s) or any obstacles within the workspace.
Fig.~\ref{fig:augmentation} demonstrates an example of the adding procedure of an augmented edge $(v_1, v_3)$ with three valid stitching tuples in $O_{v_1, v_3}$ and how $p_1$ and $p_3$ can be connected via $p_2$.
Given that the distance between adjacent isolines is set as $l$ previously, we assign a weight $w_e=l\times k$ to each $e=(u,v)\in E^\#$ with a layer difference of $k$ (i.e., $|L_u-L_v|=k$), which approximates the additional path cost incurred by any tree containing $e$. The cost of any tree $T$ is thus updated to $c(T)=\sum_{v\in V(T)} w_v+\sum_{e\in E(T)} w_e$ in the MMRTC solving.
Once the augmented edge set $E^\#$ and the corresponding stitching tuple sets $O$ are constructed, the original isograph $G$ is updated by setting $E=E\cup E^\#$, and the same MMRTC model is solved on the augmented $G$.

\begin{algorithm}[tb]
\DontPrintSemicolon
\linespread{0.6}\selectfont
\caption{MMRTC Solution Refinement}\label{alg:sol_refine}
\SetKwInput{KwInput}{Input}
\KwInput{isograph $G=(V,E)$, optimal MMRTC solution $\mathcal{T}$}
optimized solution $\mathcal{T}^*\gets \mathcal{T}$, set of used isovertices $U\gets\emptyset$\;\label{alg:sol_refine:init_1}
$M\gets$ set of all repeatedly visited isovertices in $\mathcal{T}$\;\label{alg:sol_refine:init_2}
call \textsc{AddImprovingRepetition}($\mathcal{T}, M, U$) if $M=\emptyset$\;\label{alg:sol_refine:air1}
max-heapify $M$ ordered by the number of occurrences\;\label{alg:sol_refine:sort}
\While{$M\neq\emptyset$}{\label{alg:sol_refine:loop}
    $u\gets M.pop()$\;\label{alg:sol_refine:pop}
    $\mathcal{T}_u\gets$ set of all trees containing $u$ in current solution $\mathcal{T}$\;\label{alg:sol_refine:init_Ts}
    \For{$(u,v)\in\{(u,v)\in E\,|\,v\notin U\}$}{\label{alg:sol_refine:iter_edge}
        $h, \mathcal{T}_u\gets \textsc{PairwiseIsoverticesSplitting}(\mathcal{T}_u, u, v)$\;\label{alg:sol_refine:pis}
        set $h^*$ to $h$ and $\mathcal{T}^*_u$ to $\mathcal{T}_u$ if $h < h^*$\;\label{alg:sol_refine:record}
    }
    % use $\mathcal{T}_u^*$ to update $\mathcal{T}$, $U\gets U\cup\{u, v\}$, $M\gets M/\{v\}$ if $v\in M$\;\label{alg:sol_refine:update}
    use $\mathcal{T}_u^*$ to update $\mathcal{T}$, $U\gets U\cup\{u, v\}$, $M\gets M/\{v\}$\;\label{alg:sol_refine:update}
    set $\mathcal{T}^*$ to $\mathcal{T}$ if its evaluated makespan is smaller\;\label{alg:sol_refine:update_return}
    call \textsc{AddImprovingRepetition}($\mathcal{T}, M, U$) if $M=\emptyset$\;\label{alg:sol_refine:air2}
}
\Return $\mathcal{T}^*$\;\label{alg:sol_refine:ret}
\renewcommand{\texttt}[1]{\scshape{\textsc{#1}}}
\SetKwFunction{FMain}{AddImprovingRepetition}
\SetKwProg{Fn}{Function}{:}{}
\Fn{\FMain{$\mathcal{T}, M, U$}}
{\label{alg:air:st}
$P\gets$ set of leaf isovertices  (i.e., with a degree of $1$) $u\notin U$ in the highest-cost tree in $\mathcal{T}$ that are not from PIS splitting\;\label{alg:air:init_P}
$T, u\gets$ lowest-cost $T\in\mathcal{T}$ and any $u\in P$ such that $u$ is not in $T$ but a neighbor of some $v\notin U$ in $T$\;\label{alg:air:find}
add $u$ and edge $(u,v)$ to $T$, $M\gets M\cup\{u\}$\;\label{alg:air:add_rep}
}
\renewcommand{\texttt}[1]{\scshape{\textsc{#1}}}
\SetKwFunction{FMain}{PairwiseIsoverticesSplitting}
\SetKwProg{Fn}{Function}{:}{}
\Fn{\FMain{$\mathcal{T}_u, u, v$}}
{\label{alg:pis:st}
$h^*\gets +\infty, \mathcal{T}^*_u\gets \mathcal{T}_u$\;
\For{$\mathbf{o}=(o_1,\ldots, o_{|\mathcal{T}_u|})\in O_{u,v}^{|\mathcal{T}_u|}$}{\label{alg:pis:for_loop}
    $\mathcal{T}_u' \gets$ a copy of $\mathcal{T}_u$\;\label{alg:pis:copy_T}
    split $u, v$ into $|\mathcal{T}_u|$ new isovertices $z$'s by stitching $I_u, I_v$ via $\mathbf{o}$, each assigned to a $T\in\mathcal{T}_u'$\Comment{see Figs.~\ref{fig:pairwise_splitting}-(b)(c)(d)}\;\label{alg:pis:split}
    \For{$T\in\mathcal{T}_u$}{\label{alg:pis:loop_update_edge}
        % add the assigned new isovertex $z$ to $T$\;
        \If{$v\in T$}{
            replace each edge $(u,\cdot)$ or $(v, \cdot)$ (except edge $(u,v)$) with $(z,\cdot)$ in $T$ and remove $u,v$ from $T$\;
        }
        \Else{replace each $(u,\cdot)$ with $(z,\cdot)$ in $T$ and remove $u$ from $T$\;}
        % \textcolor{red}
        % {
        % \scalebox{0.96}{add the assigned new isovertex $z$ to $T$ and remove $u$ from $T$}\;
        % replace each edge $(u,\cdot)$ with edge $(z,\cdot)$ in $T$\;
        % \If{$v\in V(T)$}{
        %     remove $v$ from $T$ and replace each $(v,\cdot)$ with $(z,\cdot)$ in $T$\;
        %     % replace each edge $(v,\cdot)$ with edge $(z,\cdot)$ in $T$
        % }
        % }
        mark each edge $(z,\cdot)$ as nonadjacent if $O_{z,\cdot}=\emptyset$\;\label{alg:pis:edge_replacing}
    }
    $h\gets$ sum of the standard deviation of the tree costs in $\mathcal{T}_u'$ and the distance corresponding to any nonadjacent edge\;\label{alg:pis:diff}
    set $h^*$ to $h$ and $\mathcal{T}^*_u$ to $\mathcal{T}_u'$ if $h<h^*$\;
}
\Return $h^*, \mathcal{T}^*_u$\;\label{alg:pis:ret}
}
\end{algorithm}

\subsection{Optimization: MMRTC Solution Refinement}\label{subsec:iso-splitting}
Despite that isograph augmentation reduces isovertex repetitions in the optimal MMRTC solution, two bottlenecks persist in achieving a better MCPP solution. 
The first bottleneck results from certain isovertex repetitions that remain unresolved by augmentation alone, notably when multiple robots share the same root isovertex or multiple trees use the same vertex.
The second bottleneck arises from the limitation of an optimal MMRTC solution in balancing tree costs when the traversing costs of the isolines vary significantly.
To tackle the above two bottlenecks, we propose the MMRTC solution refinement process (Alg.~\ref{alg:sol_refine}) that leverages two functions \textsc{PairwiseIsoverticesSplitting} (PIS) and \textsc{AddImprovingRepetition} (AIR): PIS disperses the coverage of the isoline of an isovertex with repetitions among multiple robots, while AIR introduces \textit{improving repetition} by selectively adding an isovertex from a higher-cost tree to a lower-cost tree.
Both PIS and AIR are crucial in refining the MMRTC solution: PIS directly addresses the issue of isovertex repetitions, while AIR strategically adjusts coverage load distribution to balance costs among the trees, enhancing the overall MCPP solution.

\begin{figure}[tb]
\centering
\includegraphics[width=0.9\columnwidth]{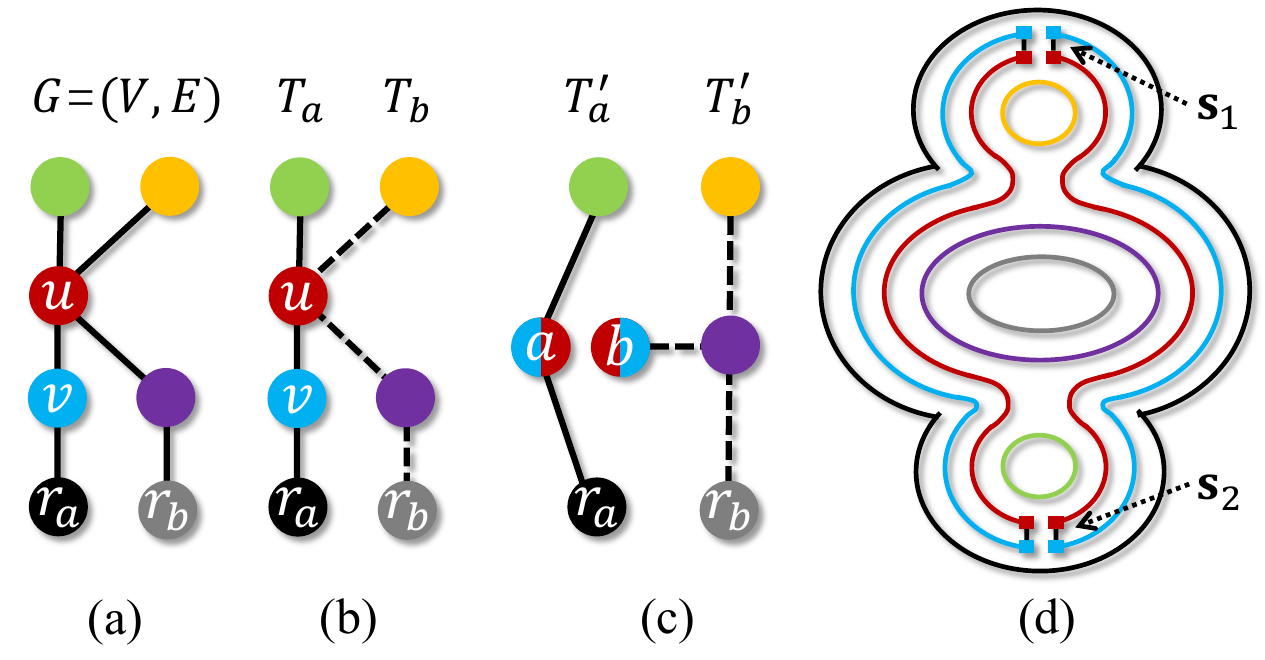}
\caption{Pairwise isovertices splitting from $u, v$ into $a, b$ at stitching tuples $o_1, o_2$. (a) Isograph $G$. (b)(c) Two trees of $G$ (in dashed and solid lines, respectively) before and after the splitting. (d) The layered isolines, each corresponding to the isovertex in the same color, and their post-split segments.}
\label{fig:pairwise_splitting}
\end{figure}

\noindent\textbf{Pseudocode:}
The MMRTC solution refinement process outlined in Alg.~\ref{alg:sol_refine} [Lines~\ref{alg:sol_refine:init_1}-\ref{alg:sol_refine:air2}] iterates through all isovertices with repetitions in decreasing order of their number of occurrences across different trees. For each such isovertex $u$, the process aims to find the best way to optimize the set $\mathcal{T}_u$ of trees containing $u$ within the MMRTC solution $\mathcal{T}$. To do so, the process calls PIS to evaluate splitting $u$ with each neighbor $v$ not used for PIS before and updates $\mathcal{T}$ to incorporate the optimized tree set $\mathcal{T}_u^*$ that yields the smallest $h$-value [Lines~\ref{alg:sol_refine:loop}-\ref{alg:sol_refine:update}], with a subsequent update to the current best solution $\mathcal{T}^*$ if $\mathcal{T}$ is better [Line~\ref{alg:sol_refine:update_return}]. The process also calls AIR to potentially add an improving repetition to an empty $M$ [Lines \ref{alg:sol_refine:air1} and \ref{alg:sol_refine:air2}].
As every iteration records isovertices used for PIS in $U$ [Line~\ref{alg:sol_refine:update}] and AIR only adds unused isovertices to $M$ [Lines~\ref{alg:air:init_P}-\ref{alg:air:find}], Alg.~\ref{alg:sol_refine} terminates after at most $|V|/2$ iterations since two new isovertices are added to $U$ on Line~\ref{alg:sol_refine:update} in each iteration.

\noindent\textbf{AIR ([Lines~\ref{alg:air:st}-\ref{alg:air:add_rep}])} identifies one leaf isovertex, unused for PIS before and not resulting from PIS splitting, from the highest-cost tree [Line~\ref{alg:air:init_P}] and adds it as an improved repetition to the lowest-cost neighboring tree [Line~\ref{alg:air:find}-\ref{alg:air:add_rep}], allowing for redistributing the tree costs.

\noindent\textbf{PIS ([Lines~\ref{alg:pis:st}-\ref{alg:pis:ret}])} takes as input not only $u$ with repetitions but also its neighbor $v$ [Line~\ref{alg:pis:st}], essential for forming a closed loop from two isoline segments (as shown in red and blue in Fig.~\ref{fig:pairwise_splitting}-(d)), and splits them into $|\mathcal{T}_u|$ new isovertices. Each new isovertex $z$ corresponds to a closed loop and is then integrated into its designated tree $T\in\mathcal{T}_u$ [Lines \ref{alg:pis:loop_update_edge}-\ref{alg:pis:edge_replacing}]. To heuristically select the best way of cost-balancing splitting, PIS evaluates each possible mapping $\mathbf{o}$ from the stitching tuples in $O_{u,v}$ to the trees in $\mathcal{T}_u$ (through the $|\mathcal{T}_u|$-th Cartesian power of $O_{u,v}$) by computing the $h$-value for its resultant tree set $\mathcal{T}_u'$ [Lines~\ref{alg:pis:for_loop}-\ref{alg:pis:diff}]. This includes: (1) Obtaining the stitching tuple set $O_{z,\cdot}$ for each new edge $(z,\cdot)$ by encompassing all valid stitching tuples on its assigned closed loop. (2) Incorporating the distance between isolines $I_z$ and $I_x$ into the $h$-value [Line~\ref{alg:pis:diff}] if an edge $(z,x)$ marked as nonadjacent (i.e., $O_{z,x}=\emptyset$) is used in the optimized solution, necessitating an additional shortest path to route between $I_z$ and $I_x$.
Fig.~\ref{fig:pairwise_splitting} demonstrates how an isovertex $u$, contained in two trees, split into two new isovertices via PIS.

\subsection{Case Study: MMRTC Solution Optimizations}
\begin{figure}[tb]
\centering
\includegraphics[width=0.98\columnwidth]{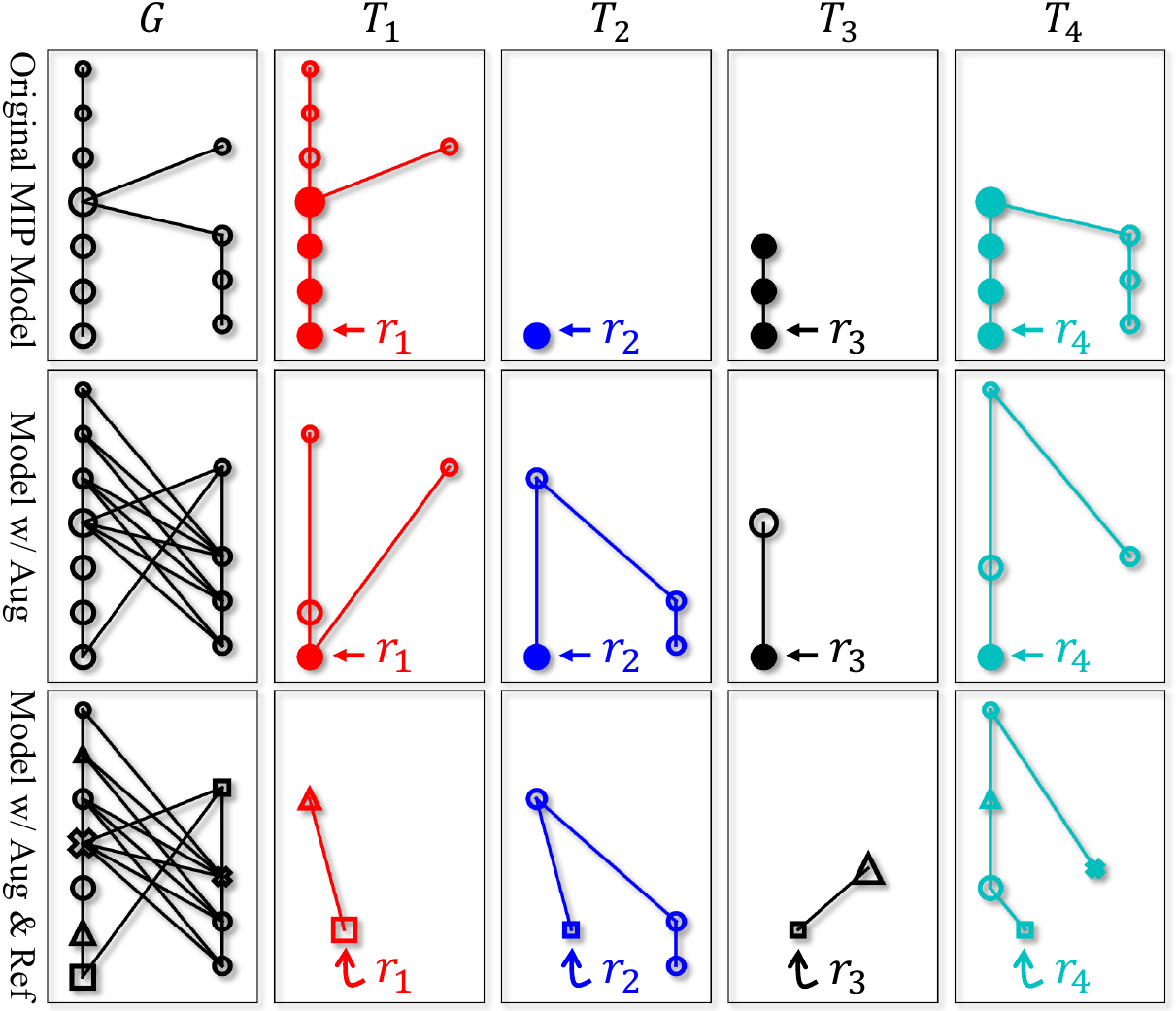}
\caption{Three MMRTC solutions $\mathcal{T}$$=$$\{T_i\}_{i=1}^4$ on isograph $G$ depicted in three rows. The weight of an isovertex corresponds to its marker size. An isovertex filled with color is covered by multiple trees. In the third row, PIS split the isovertices in the same marker (except circles) in $G$ into new isovertices in the same marker and assigned to trees in $\mathcal{T}$.}
\label{fig:case_study_opt}
\end{figure}

\begin{figure*}[tb]
\centering
\includegraphics[width=0.89\linewidth]{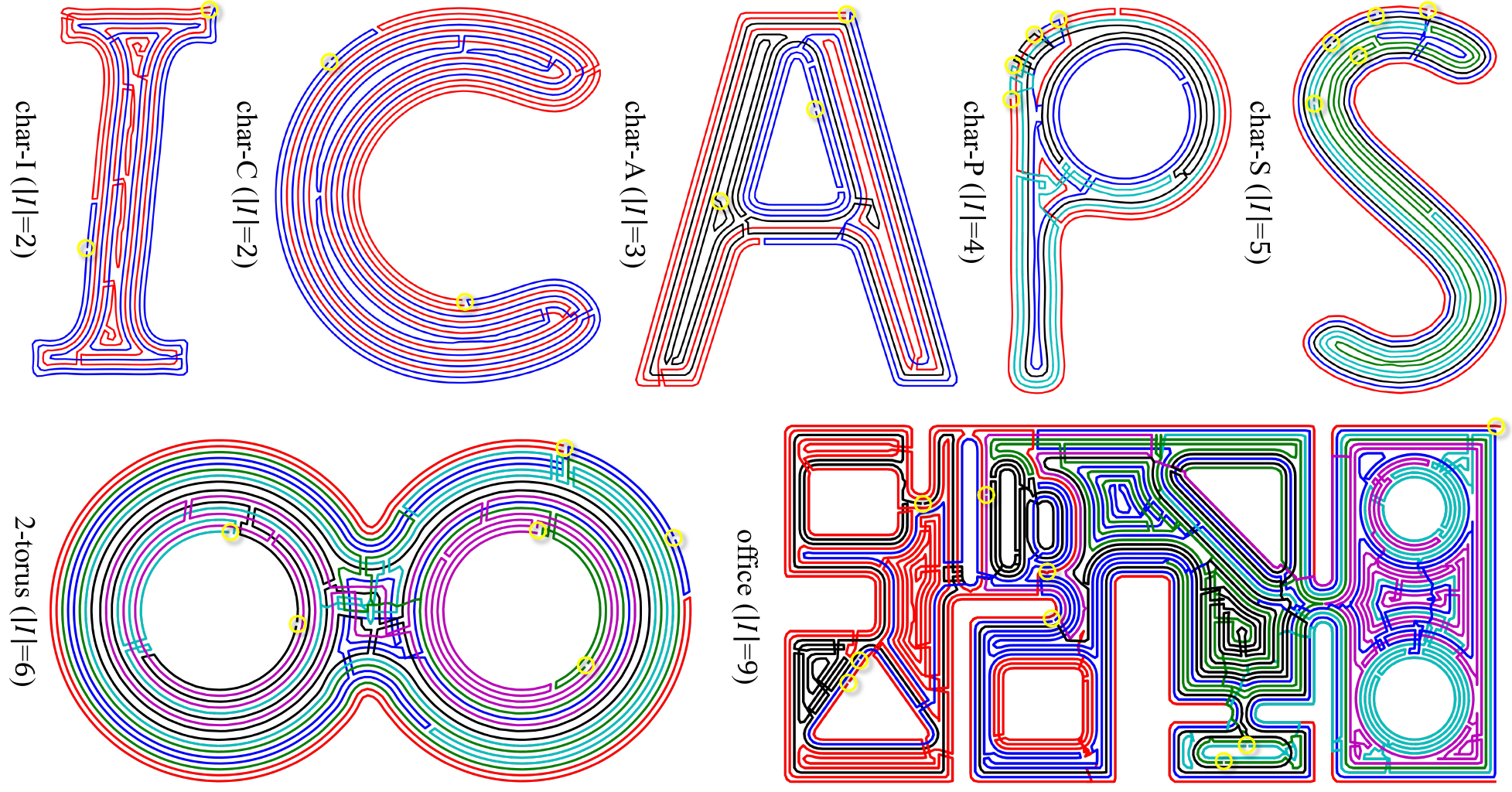}
\caption{Coverage paths from MCFS. Different paths are in different colors. Yellow circles are root positions.}
\label{fig:gallery}
\end{figure*}

We give a concrete example to better illustrate how the two aforementioned optimizations of isograph augmentation (Aug) and solution refinement (Ref) improve an MMRTC solution obtained from the original MIP model.
As shown in Fig.~\ref{fig:case_study_opt}, we use the instance \textit{char-P} of Fig.~\ref{fig:gallery}, where the four trees are rooted in the same isovertex.
The original MMRTC solution in the first row demonstrates four isovertices (filled in colors) with repetitions, yielding highly unbalanced costs among trees. With Aug ($\delta$ is set to $4$) in the second row, the sparsity of the isograph $G$ decreases and thereby provides more routing options starting at the root, making the solution less isovertex repetitions and more cost balanced. With both Aug and Ref in the third row, the solution is further improved by deduplicating all isovertices with repetitions and dynamically adjusting the costs between the trees.

\section{Empirical Evaluation}
This section presents our experimental results on a 3.49 GHz Apple® M2 CPU laptop with 16GB RAM.
% Our code will be publicly available upon acceptance of this paper.

\begin{table}[tb]
\renewcommand{\arraystretch}{0.9}
\setlength\tabcolsep{2.8pt}
\fontsize{15}{18}\selectfont\scalebox{0.59}{
\begin{tabular}{||c|c|c|c|c|c|c|c||}
\hline
\textbf{Selectors} & \textbf{\textit{char-I}} & \textbf{\textit{char-C}} & \textbf{\textit{char-A}} & \textbf{\textit{char-P}} & \textbf{\textit{char-S}} & \textbf{\textit{2-torus}} & \textbf{\textit{office}} \\\hline
random  & 2.824 &	0.924 & 1.228 & 2.095 & 1.084 & 1.070 & 12.93 \\\hline
CFS & 1.306 & 0.747 & 0.848 & 1.724 & 0.887 & 0.819 & 11.77 \\\hline
MCS & 1.269 & 0.782 & 0.874 & 1.277 & 0.960 & 0.969 & 8.289 \\\hline
\end{tabular}
}
\caption{Curvature comparison between stitching tuple selectors in the unified version of CFS for single-robot CPP.}
\label{tab:selectors}
\end{table}

\begin{table}[tb]
\centering
\renewcommand{\arraystretch}{0.9}
\setlength\tabcolsep{1.2pt}
\fontsize{15}{18}\selectfont\scalebox{0.59}{
\begin{tabular}{||c|c|c|c|c|c|c|c|c|c||} \hline 
& \multicolumn{2}{c|}{\textbf{Method}} & \textbf{\textit{char-I}} & \textbf{\textit{char-C}} & \textbf{\textit{char-A}} & \textbf{\textit{char-P}} & \textbf{\textit{char-S}} & \textbf{\textit{2-torus}} & \textbf{\textit{office}} \\ \hline
&\multicolumn{2}{c|}{robots}&2&2&3&4&5&6&9\\\hline\hline
\multirow{6}{*}{\rotatebox[origin=c]{270}{\textbf{Makespan} ($\tau$)}} & \multicolumn{2}{c|}{TMC} 
& 99.94 & 136.3 & 87.75 & 75.19 & 62.51 & 133.7 & 154.0 \\ \cline{2-10}
& \multicolumn{2}{c|}{TMSTC$^*$} 
& 91.33 & 117.9 & 84.35 & 50.63 & 56.41 & 113.9 & 238.1 \\ \cline{2-10}
& \multirow{4}{*}{\rotatebox[origin=c]{270}{MCFS}} & NONE & 132.3 & 179.8 & 75.4 & 106.8 & 50.46 & 174.2 & 291.0 \\ \cline{3-10}
& & +REF & \textbf{69.74} & 125.7 & 63.44 & 52.86 & 50.46 & 108.9 & 213.4 \\ \cline{3-10}
& & +AUG & 85.37 & 106.3 & 63.14 & 48.23 & 46.26 & 87.86 & 155.5 \\ \cline{3-10}
& & +BOTH & 70.75 & \textbf{105.0} & \textbf{63.14} & \textbf{35.13} & \textbf{36.04} & \textbf{80.73} & \textbf{141.2} \\ \hline\hline

\multirow{6}{*}{\rotatebox[origin=c]{270}{\textbf{Curvature}}} & \multicolumn{2}{c|}{TMC} 
& 2.541 & 3.433 & 7.482 & 6.115 & 5.011 & 3.341 & 8.459 \\ \cline{2-10}
& \multicolumn{2}{c|}{TMSTC$^*$} 
& 2.476 & 1.801 & 2.655 & 2.869 & 2.259 & 1.335 & 2.117 \\ \cline{2-10}
& \multirow{4}{*}{\rotatebox[origin=c]{270}{MCFS}} & NONE & 1.129 & 0.776 & \textbf{0.950} & 0.970 & 1.050 & 1.299 & 1.192 \\ \cline{3-10}
& & +REF & 2.512 & 0.842 & 0.981 & 1.184 & 1.050 & 1.357 & 1.737 \\ \cline{3-10}
& & +AUG & \textbf{0.972} & \textbf{0.758} & 1.047 & \textbf{0.828} & \textbf{0.787} & 1.070 & \textbf{1.087} \\ \cline{3-10}
& & +BOTH & 1.026 & 0.795 & 1.047 & 1.428 & 1.068 & \textbf{1.064} & 1.352 \\ \hline\hline

\multirow{6}{*}{\rotatebox[origin=c]{270}{\textbf{Coverage}}} & \multicolumn{2}{c|}{TMC} 
& 86.8\% & 87.6\% & 88.4\% & 88.0\% & 85.8\% & 91.5\% & 89.2\% \\ \cline{2-10}
& \multicolumn{2}{c|}{TMSTC$^*$} 
& 90.6\% & 92.4\% & \textbf{91.0\%} & \textbf{90.2\%} & 91.2\% & 93.7\% & \textbf{91.3\%} \\ \cline{2-10}
& \multirow{4}{*}{\rotatebox[origin=c]{270}{MCFS}} & NONE & 91.1\% & 92.4\% & 89.5\% & 89.4\% & 91.9\% & \textbf{94.6\%} & 91.2\% \\ \cline{3-10}
& & +REF & 91.1\% & 92.4\% & 89.4\% & 89.4\% & 91.9\% & 94.5\% & 91.1\% \\ \cline{3-10}
& & +AUG & \textbf{91.1\%} & \textbf{92.5\%} & 89.4\% & 89.4\% & \textbf{91.9\%} & 94.5\% & 91.1\% \\ \cline{3-10}
& & +BOTH & 91.0\% & 92.4\% & 89.4\% & 89.4\% & 91.8\% & 94.5\% & 91.1\% \\ \hline\hline

\multirow{6}{*}{\rotatebox[origin=c]{270}{\textbf{Overlapping}}} & \multicolumn{2}{c|}{TMC} 
& 8.76\% & 7.76\% & \textbf{5.59\%} & 7.89\% & 18.8\% & 15.8\% & 15.3\% \\ \cline{2-10}
& \multicolumn{2}{c|}{TMSTC$^*$} 
& 8.12\% & 6.25\% & 9.37\% & 13.1\% & 16.5\% & 15.5\% & 17.1\% \\ \cline{2-10}
& \multirow{4}{*}{\rotatebox[origin=c]{270}{MCFS}} & NONE & 82.6\% & 5.46\% & 5.91\% & \textbf{62.5\%} & 6.79\% & 86.6\% & 50.2\% \\ \cline{3-10}
& & +REF & \textbf{6.50\%} & \textbf{5.44\%} & 5.92\% & 7.95\% & \textbf{6.79\%} & 25.0\% & 24.0\% \\ \cline{3-10}
& & +AUG & 22.4\% & 6.41\% & 6.75\% & 22.2\% & 7.48\% & 20.0\% & 24.5\% \\ \cline{3-10}
& & +BOTH & 7.27\% & 6.25\% & 6.63\% & 10.8\% & 7.41\% & \textbf{9.62\%} & \textbf{13.1\%} \\ \hline\hline

\multirow{6}{*}{\rotatebox[origin=c]{270}{\textbf{Runtime}}} & \multicolumn{2}{c|}{TMC} 
& 0.25s & 1.26s & 0.97s & 0.33s & 76.0s & 30.4m & 31.2m \\ \cline{2-10}
& \multicolumn{2}{c|}{TMSTC$^*$} 
& 1.21s & 1.78s & 1.77s & 1.02s & 2.70s & 8.22s & \textbf{27.9s} \\ \cline{2-10}
& \multirow{4}{*}{\rotatebox[origin=c]{270}{MCFS}} & NONE & \textbf{0.24s} & \textbf{0.38s} & \textbf{0.44s} & \textbf{0.29s} & \textbf{0.31s} & \textbf{1.57s} & 30.1m \\ \cline{3-10}
& & +REF & 8.59s & 11.7s & 8.60s & 5.08s & 0.60s & 39.8s & 33.1m \\ \cline{3-10}
& & +AUG & 0.34s & 0.60s & 0.85s & 0.46s & 0.60s & 13.9m & 30.2m \\ \cline{3-10}
& & +BOTH & 7.13s & 12.5s & 20.0s & 7.89s & 15.6s & 15.2m & 37.5m \\ \hline

\end{tabular}}
\caption{Solution quality for different MCPP algorithms.}
\label{tab:sol_qual}
\end{table}

\noindent\textbf{Setup:}
The MMRTC MIP model for MCFS is solved using the Gurobi solver~\cite{gurobi} with a runtime limit of 30 minutes and an MST-based initial solution for warm start-up~\cite{tang2023mixed}.
Whenever MCFS is equipped with isograph augmentation, the hyperparameter $\delta$ is set to $\min\{|I|, 4\}$, where $|I|$ is the number of robots for the MCPP instance, balancing between the MMRTC model complexity and the solution quality.

\noindent\textbf{Instances:}
As existing MCPP benchmarks like~\cite{tang2023mixed} are tailored for grid-based methods on 2D grid maps, we use a more diverse set of workspaces to design MCPP instances displayed in Fig.~\ref{fig:gallery}, ranging from fully non-rectilinear (\textit{2-torus}) to mostly rectilinear (\textit{office}) ones.
% We conduct the experiments using MCPP instances displayed in Fig.~\ref{fig:gallery}, where the polygon workspace that needs to be covered is already filled with coverage paths via MCFS.
The distance $l$ between adjacent isolines in all instances is $0.1$, which is also the cover diameter of the robots.
The number of robots ($|I|$) in the instances ranges from $2$ to $9$.
In \textit{char-I} and \textit{char-P}, two robots and four robots share the same root isovertex, respectively. In \textit{2-torus}, three pairs of robots share three root isovertices, respectively. In all other instances, robots start from different root isovertices.

\noindent\textbf{Metrics:}
In addition to the makespan $\tau$, we report the following metrics to evaluate an MCPP method and its solution:
(1) Curvature: Average curvature of all paths (smaller values indicate smoother paths).
(2) Coverage: Ratio between the covered area and the total workspace.
(3) Overlapping: Ratio between the repeatedly covered area and the total workspace area.
(4) Runtime: Total runtime of the method, including the MIP model solving time (when applicable).

\noindent\textbf{Stitching Tuple Selectors:}
Tab.~\ref{tab:selectors} compares curvature among the random, CFS, and MCS stitching tuple selectors. Both CFS and MCS selectors outperform the random selector, with average reductions of 24.6\% and 27.9\%, respectively. For less complex workspaces such as \textit{ 2-torus} that can be filled with smooth isolines, the CFS selector with staircase-like stitching paths outperforms the MCS selector since the MCS selector struggles to distinguish small curvature differences. However, for complex workspaces like \textit{office}, the MCS selector significantly excels by strategically selecting sharp corner points as stitching tuples, thereby substantially reducing the curvature.  
Based on these findings, the MCS selector will be used in the MCFS framework for the remainder of our experiments.

\begin{figure}[tb]
\centering
\includegraphics[width=0.995\linewidth]{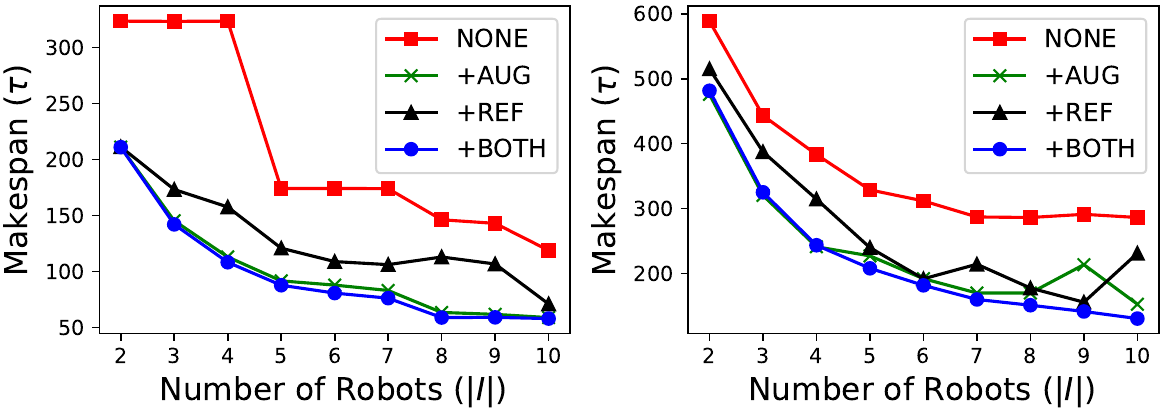}
\caption{MCFS makespan comparisons on instances \textit{2-torus} (left) and \textit{office} (right) with different number of robots.}
\label{fig:num_robots}
\end{figure}

\noindent\textbf{Ablation Study:}
To validate the effectiveness of isograph augmentation (Aug) and MMRTC solution refinement (Ref) for MCFS, Tab.~\ref{tab:sol_qual} reports results for four MCFS variants: using only the original MMRTC solution, with Aug, with Ref, and combining both (labeled \texttt{\textbf{NONE}}, \texttt{\textbf{AUG}}, \texttt{\textbf{REF}}, and \texttt{\textbf{BOTH}}, respectively).
Compared to NONE, REF and AUG reduce the makespan by an average of 29.7\% and 36.0\%, respectively.
For \textit{char-I}, \textit{char-P}, \textit{2-torus}, and \textit{office}, this reduction is attributed to decreased overlapping ratio,  particularly where the robot root positions are identical or adjacent. BOTH further enhances this effect in more complex instances for more complex instances like \textit{2-torus} and \textit{office}, doubling the reduction in the overlapping ratio, resulting in a greater makespan reduction.
For \textit{char-C}, \textit{char-A}, \textit{char-S} where overlapping ratios of NONE are already low, the makespan reduction of REF results from the iterative cost-balancing procedure, whereas the makespan reduction of AUG results from a larger MMRTC solution space via the augmented edges. Although both REF and AUG require a longer runtime, this increase in runtime is less pronounced for complex instances where the MMRTC MIP model solving dominates.
Overall, BOTH yields the largest average makespan reduction of 43.6\% compared to NONE, combining the strengths of both REF and AUG in makespan minimization at the cost of slightly longer runtime.
Fig.~\ref{fig:num_robots} further shows the evolving performance of four MCFS variants in two instances with increasing numbers of robots. It indicates that both optimizations are crucial with more robots as each robot needs to cover fewer isolines, providing a more robust MMRTC solution improvements and thereby the makespan reductions.
Aug consistently aids in reducing makespan by expanding the MMRTC solution space, though it increases the complexity and runtime of the resulting MIP model, and Ref effectively redistribute the costs of the imbalanced MMRTC trees through isovertex splitting. 
Specifically, for the \textit{2-torus} instance, Aug plays a pivotal role, whereas Ref contributes only marginal improvements; for the \textit{office} instance, either Aug or Ref individually contributes significantly to the solution improvement, while the combined use of both demonstrates a more robust enhancement in the solution.
% The figure also shows that for instances with large number of robots, the MIP models become too complex for all four variants to obtain satisfactory MMRTC solutions within the runtime limit, whereas Ref, employed by REF and BOTH, continues to significantly reduce the makespan of the resulting suboptimal MMRTC solutions.

% \noindent\textbf{Different Numbers of Robots and Unique Roots:}
% Fig.~\ref{fig:num_robots} further shows the evolving performance of four MCFS variants for \textit{office} with increasing numbers of robots and unique roots. Fig.~\ref{fig:num_robots} (Left) indicates that Ref is more crucial with more robots because each robot needs to cover fewer isolines, often leading to imbalanced MMRTC trees, making isovertex splitting more effective in cost balancing. Aug consistently aids in reducing makespan by expanding the MMRTC solution space, though it increases the complexity and runtime of the resulting MIP model. The figure also shows that for $|I|\geq 7$, the MIP models become too complex for all four variants to obtain satisfactory MMRTC solutions within the runtime limit, whereas Ref, employed by REF and BOTH, continues to significantly reduce the makespan of the resulting suboptimal MMRTC solutions.

\begin{figure}[tb]
\centering
\includegraphics[width=0.735\linewidth]{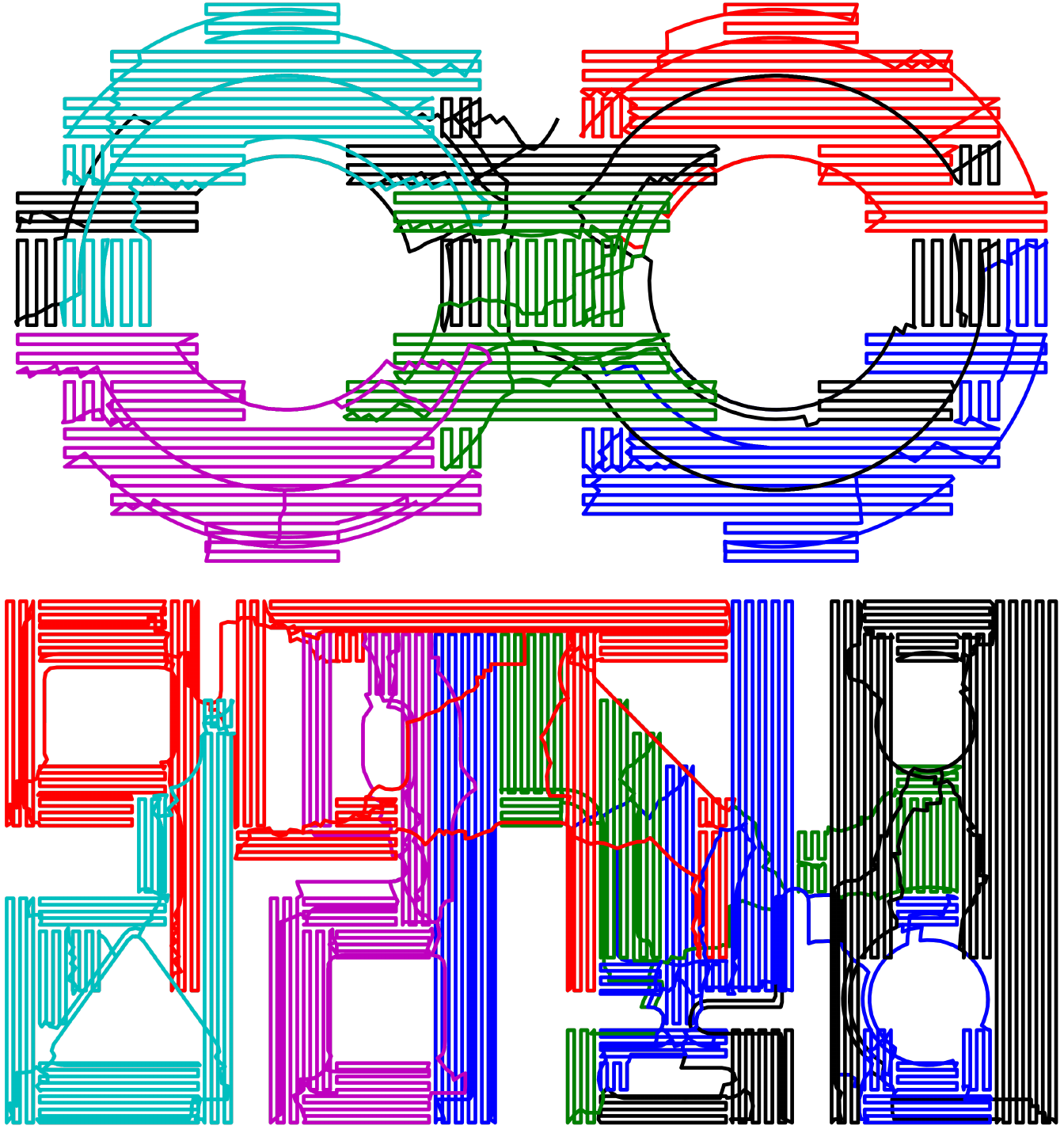}
\caption{TMC MCPP solutions for \textit{2-torus} and \textit{office}.}
\label{fig:TMC_demo}
\end{figure}

\begin{figure}[tb]
\centering
\includegraphics[width=0.735\linewidth]{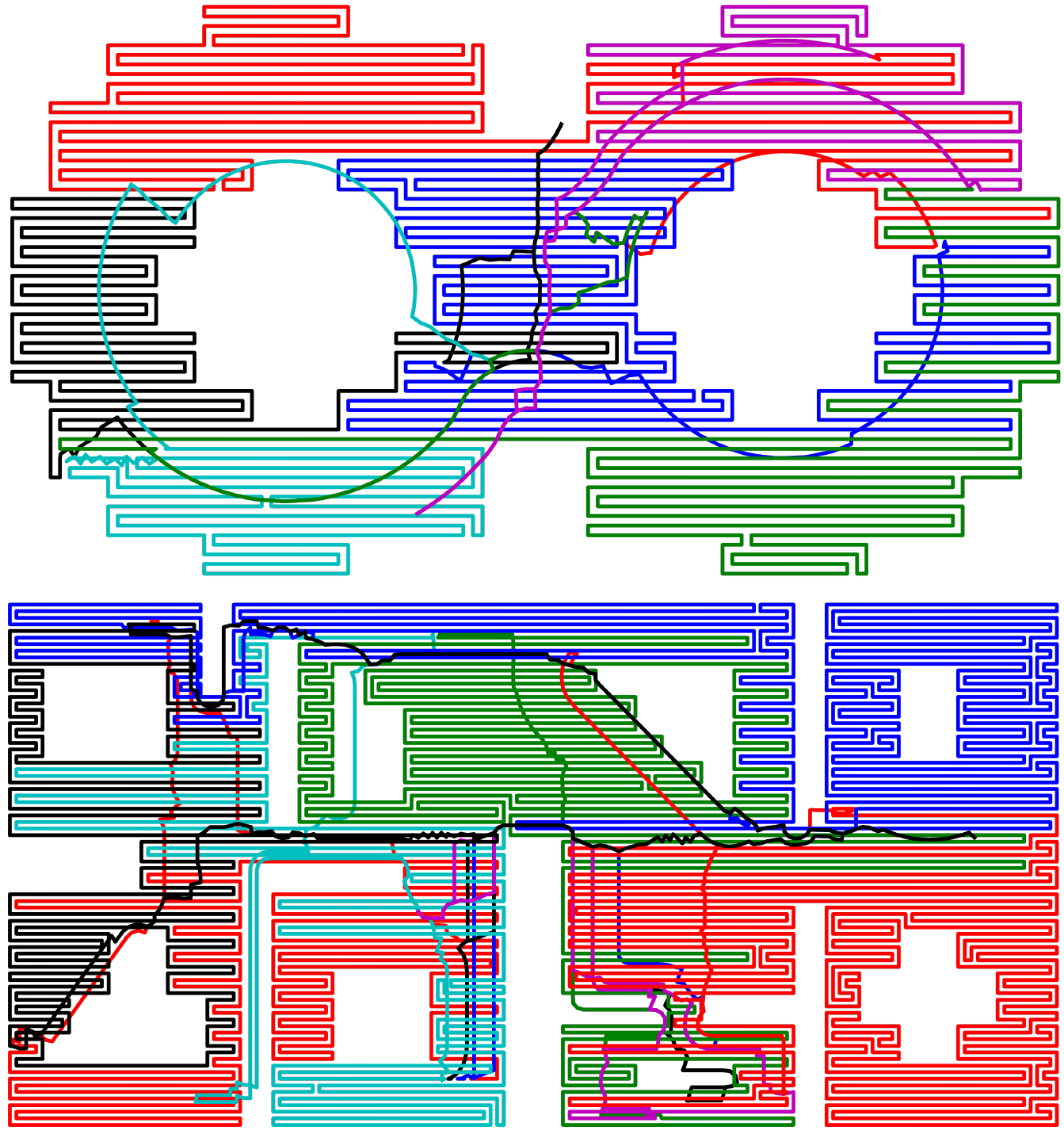}
\caption{TMSTC$^*$ MCPP solutions for \textit{2-torus} and \textit{office}.}
\label{fig:TMSTCStar_demo}
\end{figure}

\noindent\textbf{Comparison:}
We compare MCFS (+BOTH) with two state-of-the-art grid-based MCPP methods, TMC~\cite{vandermeulen2019turn} and TMSTC$^*$~\cite{lu2023tmstc}, that minimize path turns.
To adapt TMC and TMSTC$^*$ to the non-rectilinear workspaces in our instances, we use overlay grids to approximate the workspaces, followed by shortest pathfinding for robot return to root positions post-coverage. Note that the reported coverage and overlapping ratios for TMC and TMSTC$^*$ are approximations due to the workspace approximation and small intersection of their coverage paths with obstacles, whereas the values for MCFS are exact.
In Tab.~\ref{tab:sol_qual}, while the average coverage ratios of TMC, TMSTC$^*$, and MCFS are comparably close (with a 3.51\% variance), MCFS demonstrates an average makespan reduction of 32.0\% and 27.9\%, curvature reduction of 75.7\% and 47.8\%, and overlapping ratio reduction of 13.6\% and 20.9\% compared to TMC and TMSTC$^*$, respectively. Both MCFS and TMC require longer runtime due to solving MIP models for MMRTC and MTSP, respectively, especially in instances with larger isographs or more robots (e.g., \textit{office}).
Fig.~\ref{fig:TMC_demo} and Fig.~\ref{fig:TMSTCStar_demo} visualize the coverage paths via TMC and TMSTC$^*$, respectively. These paths exhibit a back-and-forth boustrophedon pattern, leading to high curvature and imperfect coverage around complex obstacles. In contrast, MCFS notably excels in generating smooth paths that efficiently contour around arbitrarily shaped obstacles, a clear visual advantage over the other methods as shown in Fig.~\ref{fig:gallery}.

\section{Conclusions}
We proposed the MCFS framework, an innovative approach that blends principles from computer graphics and automated planning to tackle the challenges of covering arbitrarily shaped workspaces in complex MCPP tasks.
%MCFS leverages our novel unified version of CFS to bring scalability and versatility for multi-robot scenarios by computing multiple rooted trees that jointly cover an input graph of isolines. We also developed two effective optimization techniques that significantly enhance the solution quality. We validated the effectiveness of MCFS in various scenarios through rigorous experimentation and analyses, benchmarked against state-of-the-art MCPP methods.
Future work includes improving isoline quality to further boost the coverage ratio,
as the coverage of MCFS highly depends on whether the generated isolines can adequately cover the workspace.
incorporating kinodynamic constraints into the generation and stitching procedures of the isolines, and developing heuristics to accelerate the PIS function and the MMRTC solving for large numbers of robots or isolines.

\section*{Acknowledgements}
This work was supported by the NSERC under grant number RGPIN2020-06540 and a CFI JELF award.

\bibliography{ref}

\begin{thebibliography}{31}
\providecommand{\natexlab}[1]{#1}

\bibitem[{Acar et~al.(2002)Acar, Choset, Rizzi, Atkar, and Hull}]{acar2002morse}
Acar, E.~U.; Choset, H.; Rizzi, A.~A.; Atkar, P.~N.; and Hull, D. 2002.
\newblock Morse decompositions for coverage tasks.
\newblock \emph{The International Journal of Robotics Research}, 21(4): 331--344.

\bibitem[{Almadhoun et~al.(2019)Almadhoun, Taha, Seneviratne, and Zweiri}]{almadhoun2019survey}
Almadhoun, R.; Taha, T.; Seneviratne, L.; and Zweiri, Y. 2019.
\newblock A survey on multi-robot coverage path planning for model reconstruction and mapping.
\newblock \emph{SN Applied Sciences}, 1: 1--24.

\bibitem[{Bochkarev and Smith(2016)}]{bochkarev2016minimizing}
Bochkarev, S.; and Smith, S.~L. 2016.
\newblock On minimizing turns in robot coverage path planning.
\newblock In \emph{CASE}, 1237--1242.

\bibitem[{Choset(2000)}]{choset2000coverage}
Choset, H. 2000.
\newblock Coverage of known spaces: The boustrophedon cellular decomposition.
\newblock \emph{Autonomous Robots}, 9: 247--253.

\bibitem[{Collins et~al.(2021)Collins, Ghassemi, Esfahani, Doermann, Dantu, and Chowdhury}]{collins2021scalable}
Collins, L.; Ghassemi, P.; Esfahani, E.~T.; Doermann, D.; Dantu, K.; and Chowdhury, S. 2021.
\newblock Scalable coverage path planning of multi-robot teams for monitoring non-convex areas.
\newblock In \emph{ICRA}, 7393--7399.

\bibitem[{Even et~al.(2004)Even, Garg, K{\"o}nemann, Ravi, and Sinha}]{even2004min}
Even, G.; Garg, N.; K{\"o}nemann, J.; Ravi, R.; and Sinha, A. 2004.
\newblock Min--max tree covers of graphs.
\newblock \emph{Operations Research Letters}, 32(4): 309--315.

\bibitem[{Gabriely and Rimon(2001)}]{gabriely2001spanning}
Gabriely, Y.; and Rimon, E. 2001.
\newblock Spanning-tree based coverage of continuous areas by a mobile robot.
\newblock \emph{Annals of Mathematics and Artificial Intelligence}, 31: 77--98.

\bibitem[{Gibson et~al.(2021)Gibson, Rosen, Stucker, Khorasani, Rosen, Stucker, and Khorasani}]{gibson2021additive}
Gibson, I.; Rosen, D.~W.; Stucker, B.; Khorasani, M.; Rosen, D.; Stucker, B.; and Khorasani, M. 2021.
\newblock \emph{Additive manufacturing technologies}, volume~17.
\newblock Springer.

\bibitem[{{Gurobi Optimization, LLC}(2023)}]{gurobi}
{Gurobi Optimization, LLC}. 2023.
\newblock {Gurobi Optimizer Reference Manual}.

\bibitem[{Hazon and Kaminka(2005)}]{hazon2005redundancy}
Hazon, N.; and Kaminka, G.~A. 2005.
\newblock Redundancy, efficiency and robustness in multi-robot coverage.
\newblock In \emph{ICRA}, 735--741.

\bibitem[{Kapoutsis, Chatzichristofis, and Kosmatopoulos(2017)}]{kapoutsis2017darp}
Kapoutsis, A.~C.; Chatzichristofis, S.~A.; and Kosmatopoulos, E.~B. 2017.
\newblock DARP: divide areas algorithm for optimal multi-robot coverage path planning.
\newblock \emph{Journal of Intelligent \& Robotic Systems}, 86: 663--680.

\bibitem[{Karapetyan et~al.(2017)Karapetyan, Benson, McKinney, Taslakian, and Rekleitis}]{karapetyan2017efficient}
Karapetyan, N.; Benson, K.; McKinney, C.; Taslakian, P.; and Rekleitis, I. 2017.
\newblock Efficient multi-robot coverage of a known environment.
\newblock In \emph{IROS}, 1846--1852.

\bibitem[{Latombe and Latombe(1991)}]{latombe1991exact}
Latombe, J.-C.; and Latombe, J.-C. 1991.
\newblock Exact cell decomposition.
\newblock \emph{Robot Motion Planning}, 200--247.

\bibitem[{Lockwood(1967)}]{lockwood1967book}
Lockwood, E.~H. 1967.
\newblock \emph{A book of curves}.
\newblock Cambridge University Press.

\bibitem[{Lu et~al.(2023)Lu, Zeng, Tang, Lam, and Wen}]{lu2023tmstc}
Lu, J.; Zeng, B.; Tang, J.; Lam, T.~L.; and Wen, J. 2023.
\newblock TMSTC*: A Path Planning Algorithm for Minimizing Turns in Multi-robot Coverage.
\newblock \emph{IEEE Robotics and Automation Letters}, 8(8): 5275--5282.

\bibitem[{Mannadiar and Rekleitis(2010)}]{mannadiar2010optimal}
Mannadiar, R.; and Rekleitis, I. 2010.
\newblock Optimal coverage of a known arbitrary environment.
\newblock In \emph{ICRA}, 5525--5530.

\bibitem[{Maple(2003)}]{maple2003geometric}
Maple, C. 2003.
\newblock Geometric design and space planning using the marching squares and marching cube algorithms.
\newblock In \emph{2003 international conference on geometric modeling and graphics, 2003. Proceedings}, 90--95. IEEE.

\bibitem[{Oksanen and Visala(2009)}]{oksanen2009coverage}
Oksanen, T.; and Visala, A. 2009.
\newblock Coverage path planning algorithms for agricultural field machines.
\newblock \emph{Journal of Field Robotics}, 26(8): 651--668.

\bibitem[{Rekleitis et~al.(2008)Rekleitis, New, Rankin, and Choset}]{rekleitis2008efficient}
Rekleitis, I.; New, A.~P.; Rankin, E.~S.; and Choset, H. 2008.
\newblock Efficient boustrophedon multi-robot coverage: an algorithmic approach.
\newblock \emph{Annals of Mathematics and Artificial Intelligence}, 52: 109--142.

\bibitem[{Ren, Sun, and Guo(2009)}]{ren2009combined}
Ren, F.; Sun, Y.; and Guo, D. 2009.
\newblock Combined reparameterization-based spiral toolpath generation for five-axis sculptured surface machining.
\newblock \emph{International Journal of Advanced Manufacturing Technology}, 40: 760--768.

\bibitem[{Song et~al.(2022)Song, Yu, Qiu, Sun, Lang, Luo, Shen, and Wang}]{song2022multi}
Song, H.; Yu, J.; Qiu, J.; Sun, Z.; Lang, K.; Luo, Q.; Shen, Y.; and Wang, Y. 2022.
\newblock Multi-UAV Disaster Environment Coverage Planning with Limited-Endurance.
\newblock In \emph{ICRA}, 10760--10766.

\bibitem[{Tang and Ma(2023)}]{tang2023mixed}
Tang, J.; and Ma, H. 2023.
\newblock Mixed Integer Programming for Time-Optimal Multi-Robot Coverage Path Planning with Heuristics.
\newblock \emph{IEEE Robotics and Automation Letters}, 8(10): 6491--6498.

\bibitem[{Tang and Ma(2024)}]{tang2024large}
Tang, J.; and Ma, H. 2024.
\newblock Large-Scale Multi-Robot Coverage Path Planning via Local Search.
\newblock In \emph{Proceedings of the AAAI Conference on Artificial Intelligence}, volume~38, 17567--17574.

\bibitem[{Tang, Sun, and Zhang(2021)}]{tang2021mstc}
Tang, J.; Sun, C.; and Zhang, X. 2021.
\newblock {MSTC}$^*$: Multi-robot Coverage Path Planning under Physical Constrain.
\newblock In \emph{ICRA}, 2518--2524.

\bibitem[{Tomaszewski(2020)}]{Tomaszewski-2020-125840}
Tomaszewski, C.~K. 2020.
\newblock \emph{Constraint-Based Coverage Path Planning: A Novel Approach to Achieving Energy-Efficient Coverage}.
\newblock Ph.D. thesis, Carnegie Mellon University, Pittsburgh, PA.

\bibitem[{Vandermeulen, Gro{\ss}, and Kolling(2019)}]{vandermeulen2019turn}
Vandermeulen, I.; Gro{\ss}, R.; and Kolling, A. 2019.
\newblock Turn-minimizing multirobot coverage.
\newblock In \emph{ICRA}, 1014--1020.

\bibitem[{Wong and MacDonald(2003)}]{wong2003topological}
Wong, S.~C.; and MacDonald, B.~A. 2003.
\newblock A topological coverage algorithm for mobile robots.
\newblock In \emph{IROS}, 1685--1690.

\bibitem[{Wu et~al.(2019)Wu, Dai, Gong, Liu, Wang, Gu, and Wang}]{wu2019energy}
Wu, C.; Dai, C.; Gong, X.; Liu, Y.-J.; Wang, J.; Gu, X.~D.; and Wang, C.~C. 2019.
\newblock Energy-efficient coverage path planning for general terrain surfaces.
\newblock \emph{IEEE Robotics and Automation Letters}, 4(3): 2584--2591.

\bibitem[{Yang et~al.(2002)Yang, Loh, Fuh, and Wang}]{yang2002equidistant}
Yang, Y.; Loh, H.~T.; Fuh, J.; and Wang, Y. 2002.
\newblock Equidistant path generation for improving scanning efficiency in layered manufacturing.
\newblock \emph{Rapid Prototyping Journal}, 8(1): 30--37.

\bibitem[{Zhao et~al.(2016)Zhao, Gu, Huang, Garcia, Chen, Tu, Benes, Zhang, Cohen-Or, and Chen}]{zhao2016connected}
Zhao, H.; Gu, F.; Huang, Q.-X.; Garcia, J.; Chen, Y.; Tu, C.; Benes, B.; Zhang, H.; Cohen-Or, D.; and Chen, B. 2016.
\newblock Connected fermat spirals for layered fabrication.
\newblock \emph{ACM Transactions on Graphics}, 35(4): 1--10.

\bibitem[{Zheng et~al.(2010)Zheng, Koenig, Kempe, and Jain}]{zheng2010multirobot}
Zheng, X.; Koenig, S.; Kempe, D.; and Jain, S. 2010.
\newblock Multirobot forest coverage for weighted and unweighted terrain.
\newblock \emph{IEEE Transactions on Robotics}, 26(6): 1018--1031.

\end{thebibliography}

\newpage
\begin{appendices}
\section{Connected Fermat Spiral (CFS)}

We aim to interpret the most relevant fundamentals in the original CFS work~\cite{zhao2016connected} using the same notation and terminologies that are consistent with this paper. Given a workspace represented as a set of boundary polylines, CFS essentially comprises three steps to convert a set $\mathcal{I}$ of space-filling contouring isolines of the workspace into a connected Fermat spiral, as detailed below.

\subsection{Identifying the Pocket Regions}
The first step identifies multiple regions of \textit{pockets}, each defined as a set of consecutive isolines in adjacent layers with a single local minimum and a single local maximum of their distances to the boundaries (e.g., Fig.~\ref{fig:fermat_spiral}-(a)).
As in the main text, we adhere to the same isograph notation $G=(V,E)$ to represent the set $\mathcal{I}$ of isolines. Each isovertex $v\in V$ is associated with a unique isoline. The edge set $E$ contains any $(u,v)$ with a nonempty connecting segment set $O_{u\rightarrow v}$ (see Eqn.~(\ref{eqn:O_u2v})), which has an edge weight of $|O_{u\rightarrow v}|$.
By computing the minimum spanning tree (MST) $M$ of $G$ (e.g., Fig.~\ref{fig:fermat_spiral}-(e)), the identification of pockets can then be easily accomplished by identifying a set $V_p\subseteq V$ of isovertices with degrees not greater than $2$ in $M$.
The set $V_p$ is then partitioned into mutually exclusive subsets, each being a connected component of $G$ and thus forming a pocket (e.g., $R_0$ to $R_4$ in Fig.~\ref{fig:fermat_spiral}-(d)).

\subsection{Generating Fermat Spiral in Pocket}
We describe the second step for converting the isolines of each pocket $P=\{v_i\}_{i=1}^k\subseteq V_p$ into a Fermat spiral.
Each pair of $v_{i}$ and $v_{i+1}$ are associated with two isolines in adjacent layers, and $v_1$ and $v_k$ are the local minima and maxima of $P$, respectively.
We denote the nearest point in $I_v$ to $\mathbf{p}$ as $\mathcal{C}_v(\mathbf{p})$ and the preceding point of $\mathbf{p}$ along $I_v$ as $\mathcal{B}_v(\mathbf{p})$.
Given the entry point $\mathbf{p}_1$ and the exit point $\mathbf{q}_1=\mathcal{B}(\mathbf{p}_1)$ of the Fermat spiral on the isoline $I_{v_1}$, for each $v_i$ with $i>1$, we obtain two unique points $\mathbf{p}_{i}=\mathcal{C}_{v_{i}}\left(\mathcal{B}_{v_{i-1}}\left(\mathbf{p}_{i-1}\right)\right)$ and $\mathbf{q}_{i}=\mathcal{C}_{v_{i}}\left(\mathcal{B}_{v_{i-1}}\left(\mathbf{q}_{i-1}\right)\right)$ on $I_{v_i}$.
Subsequently, each isoline $I_{v_i}$ is partitioned into two segments, denoted as $I_{v_i}^{A}$ and $I_{v_i}^{B}$, representing the portions from $\mathbf{p}_i$ to $\mathbf{q}_i$ and from $\mathbf{q}_i$ to $\mathbf{p}_i$ in clockwise order, respectively. If $i$ is even, we swap its corresponding $I_{v_i}^{A}$ with $I_{v_i}^{B}$. Finally, we can obtain the Fermat spiral by stitching two spirals: one routes into $v_k$ by stitching $I_{v_{i-1}}^{A}$ to $I_{v_{i}}^{A}$ for each $i=2, \ldots, k$, and the other routes out of $v_k$ by stitching each $I_{v_{i}}^{B}$ to $I_{v_{i-1}}^{B}$ for each $i=k, \ldots, 2$. Fig.~\ref{fig:fermat_spiral}-(c) shows an example of the Fermat spiral in a pocket.

\subsection{Connecting Separated Fermat Spirals}
\begin{figure}[tb]
\centering
\includegraphics[width=\columnwidth]{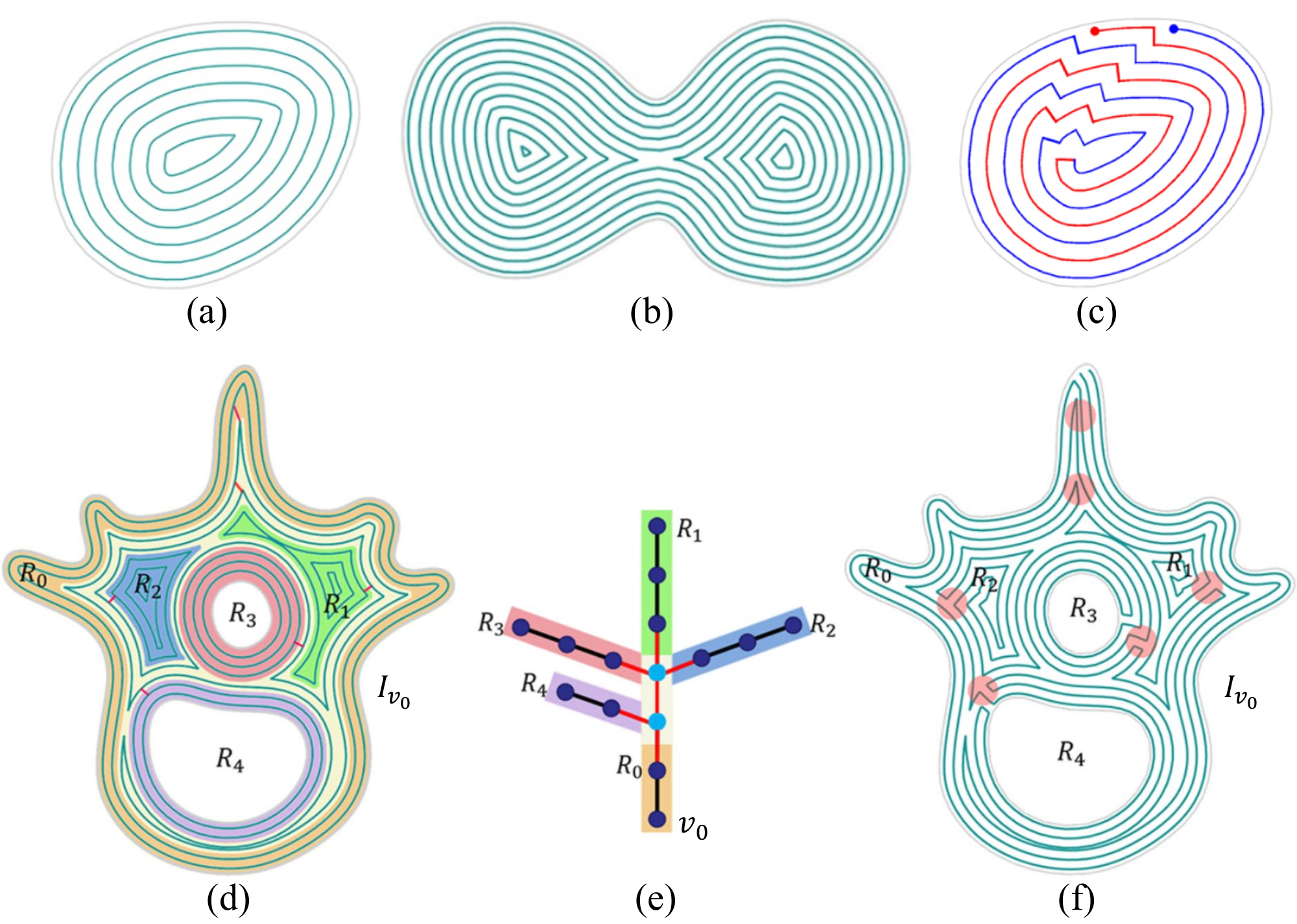}
\caption{Demonstrations adapted from the original CFS work~\cite{zhao2016connected}. (a) A pocket region. (b) A non-pocket region with two local maxima. (c) The Fermat spiral of the pocket in (a), comprised of the inward spiral (red) and the outward spiral (blue). (d) An example workspace with one exterior boundary and two interior boundaries. (e) The MST of the isograph of workspace (d). (f) The resulting connected Fermat spiral of workspace (d).}
\label{fig:fermat_spiral}
\end{figure}

The third step connects the separated Fermat spirals in the pockets by traversing the MST $M$ in a bottom-up fashion. Specifically, it starts from the isovertex $v$, where $I_v$ must be a boundary isoline containing the entry and exit points of the final connected Fermat spiral. Recall that we have grouped a set $V_p\subseteq V$ for generating the pockets.
Each pocket $P=\{v_i\}_{i=1}^k$ with its Fermat spiral and the two adjacent entry point $\mathbf{p}$ and exit point $\mathbf{q}$ are connected to the isoline $I_v$ of a vertex $v\in V/V_p$ that is a neighbor of $v_1$ or $v_k$. This connection is established by stitching $\mathbf{p}$ to $\mathcal{C}_v(\mathbf{p})$ and $\mathbf{q}$ to $\mathcal{C}_v(\mathbf{q})$. 
It is noteworthy that, according to the definition of pockets and how they are identified by the set $V_p$ on the MST $M$, the residual set $V/V_p$ will always form a connected component of $M$. Additionally, every pocket $P=\{v_i\}_{i=1}^k$ will either be rooted at $v_1$ and connected to some $v\in V/V_p$ via $v_k$, or it will be rooted at $v_k$ and connected to some $v\in V/V_p$ via $v_1$, which guarantees that the above procedure always produces a connected Fermat spiral. Figure~\ref{fig:fermat_spiral}-(f) illustrates the connected Fermat spiral of an example workspace.

\section{The MIP Model for MMRTC}
We detail the MIP model for the MMRTC problem in~\cite{tang2023mixed}, with the only difference in the definition of tree costs. Recall that in an MMRTC instance, we have a graph $G=(V,E)$, a set $I=\{1,2,...,k\}$ of indices, and a set $R=\{r_i\}_{i\in I}$ of root vertices.
The objective of MMRTC is to find a set of $k$ trees $\{T_{i}\}_{i\in I}$ minimizing the objective in Eqn.~(\ref{eqn:mmrtc}), such that each $T_i$ must be rooted at $r_i\in R$ and each vertex $v\in V$ is included in at least one tree. Compared with the original MIP model in~\citet{tang2023mixed} that defines the cost of each tree as the summation of the edge weights in Eqn.~(\ref{cstr:makespan}), here it is replaced by the summation of the each vertex weight $w_v$ to cope with our formulation for multi-robot CFS coverage.

We introduce two sets of binary variables $\textbf{x} = \{x^i_{e}\}_{e\in E}^{i\in I}$ and $ \textbf{y} = \{y^i_{v}\}_{v\in V}^{i\in I}$, where $x^i_{e}$ and $y^i_v$ take value $1$ if edge $e$ or vertex $v$ is included in the $i$-th tree $T_i$, respectively, and $0$ otherwise.
Assuming each edge has one unit of flow, we further introduce a set of non-negative continuous flow variables $\textbf{f} = \{f^i_{e,u}, f^i_{e,v}\}_{e\in E}^{i\in I}$ to represent the amount of flow assigned to vertices $u$ and $v$ for each edge $e=(u,v)\in E$.
Let $\tau$ denote the makespan and $e\sim v$ denote that $v$ is one of the endpoints of $e$. The MIP model is formulated as:
{\small
  \setlength{\abovedisplayskip}{6pt}
  \setlength{\belowdisplayskip}{\abovedisplayskip}
  \setlength{\abovedisplayshortskip}{0pt}
  \setlength{\belowdisplayshortskip}{3pt}
    \begin{align}
    \label{eqn:obj}\textbf{(MIP)}&\quad\displaystyle{\minimize_{\textbf{x}, \textbf{y}, \textbf{f}, \tau}\quad\tau} &\\
    \label{cstr:makespan}\text{s.t.}\quad&\sum_{v \in V} w_{v} y_{v}^i \leq \tau, &\forall i\in I \\
    \label{cstr:cover}&\sum_{i\in I} y_{v}^i\geq 1, &\forall v \in V\\
    \label{cstr:rooted}&y^i_{r_i} = 1, &\forall i\in I\\
    \label{cstr:tree_def}&\sum_{v\in V} y_{v}^i = 1+\sum_{e\in E} x_{e}^i, &\forall i\in I\\
    \label{cstr:acyclic_1} \quad\quad&f_{e,u}^i+f_{e,v}^i=x_{e}^i, &\forall e=(u,v)\in E, \forall i\in I\\
    \label{cstr:acyclic_2} \quad\quad&\sum_{\substack{e\in E\\ e\sim v}}f_{e,v}^i\leq 1- \frac{1}{|V|}, &\forall v\in V, \forall i\in I
    \end{align}
    \begin{flalign}
    \label{cstr:y_def} \hspace{25pt} x^i_{e}\leq y^i_v,  \quad\quad\quad\quad&\forall v\in V, \forall e\in E, e\sim v, \forall i\in I
    \end{flalign}
    \begin{flalign}
    \hspace{25pt} x^i_{e}, y^i_v\in\{0, 1\},  \quad\quad\quad\quad&\forall v\in V, \forall e\in E, \forall i\in I
    \end{flalign}
    \begin{flalign}
    \hspace{25pt} f^i_{e,u}, f^i_{e,v}, \tau\in\mathbb{R}^+, \;\;\quad\quad&\forall e=(u, v)\in E, \forall i\in I
    \end{flalign}
}%

~\citet{tang2023mixed} has established that any solution of our MIP model is feasible for its corresponding MMRTC instance, which ensures the correctness of our MIP model. The constraints of the above model can be grouped as follows:
\begin{enumerate}
\item \textit{Makespan}:
Eqn.~(\ref{cstr:makespan}) ensures that $\tau$ equals the maximum weight among all the trees, which is minimized in the objective function defined in Eqn.~(\ref{eqn:obj});
\item \textit{Cover}:
Eqn.~(\ref{cstr:cover}) enforces that each $v\in V$ is included in at least one tree;
\item \textit{Rooted}: 
Eqn.~(\ref{cstr:rooted}) enforces each $T_i$ is rooted at $r_i\in R$;
\item \textit{Tree}:
Eqn.~(\ref{cstr:tree_def}) ensures that each $T_i$ is either a single tree or a forest with cycles in some of its trees, while
Eqn.~(\ref{cstr:acyclic_1}) and (\ref{cstr:acyclic_2}) eliminate any cycles in $T_i$. Together, these constraints ensure that any tree is a single tree.
\end{enumerate}

\end{appendices}

\end{document}